%% file: paper.tex
\documentclass{adobe_research}
\usepackage[T1]{fontenc}
\usepackage{lmodern}
\usepackage{multirow}
\usepackage{soul}
\usepackage{pifont}
\usepackage[T1]{fontenc}

\usepackage{graphicx}
\usepackage{amsmath,amssymb}

\usepackage{makecell}
\usepackage[dvipsnames]{xcolor}
\usepackage{color, colortbl}
\usepackage{caption}
\usepackage{algorithm}

\input{preamble}
\usepackage{wrapfig}
\usepackage{newtxtext}

\usepackage{algpseudocode}
\definecolor{catgray}{gray}{0.92}

\usepackage[dvipsnames]{xcolor} 
\definecolor{FutureOrange}{HTML}{EC866D}

\usepackage{enumitem}

\renewcommand{\cite}{\citep}

\usepackage{tikz}

\newcommand{\ours}{\texttt{PS-VAE}\xspace}
\newcommand{\supp}{Supplementary Material}

\title{Both Semantics and Reconstruction Matter: Making Representation Encoders Ready for Text-to-Image Generation and Editing}
\author{
Shilong Zhang$^{1}$
He Zhang$^{2}$ 
Zhifei Zhang$^{2}$ 
Chongjian Ge$^{2}$ 
Shuchen Xue$^{3}$ \\
Shaoteng Liu$^{2}$ 
Mengwei Ren$^{2}$ 
Soo Ye Kim$^{2}$ 
Yuqian Zhou$^{2}$ 
Qing Liu$^{2}$ 
Daniil Pakhomov$^{2}$ \\
Kai Zhang$^{2}$ 
Zhe Lin$^{2}$ 
Ping Luo$^{1}$ \\
\vspace{0.4em}
{\small
$^{1}$The University of Hong Kong \quad
$^{2}$Adobe Research \quad
$^{3}$University of Chinese Academy of Sciences
}
}

\vspace{-5mm}
\abstract{
\input{sec/0_abstract}
}

\date{December 20th, 2025}
\adobedata[Project Page]{\url{https://jshilong.github.io/PS-VAE-PAGE/}}

\begin{document}
\maketitle

\input{sec/1_intro}
\input{sec/2_related_work}

\input{sec/3_method}
\input{sec/4_exp}
\input{sec/5_ablation}

\input{sec/6_conclusion}

\clearpage
\newpage
\bibliographystyle{assets/plainnat}
\bibliography{paper}

\end{document}

%% file: preamble.tex
\usepackage{xspace}
\makeatletter
\DeclareRobustCommand\onedot{\futurelet\@let@token\@onedot}
\def\@onedot{\ifx\@let@token.\else.\null\fi\xspace}

\def\eg{\emph{e.g}\onedot}

 \def\vs{\emph{vs}\onedot}

\makeatother

%% file: sec/0_abstract.tex
\begin{abstract}

Modern Latent Diffusion Models (LDMs) typically operate in low-level Variational Autoencoder (VAE) latent spaces that are primarily optimized for pixel-level reconstruction. To unify vision generation and understanding, a burgeoning trend is to adopt high-dimensional features from representation encoders as generative latents. However, we empirically identify two fundamental obstacles in this paradigm: (1) the discriminative feature space lacks compact regularization, making diffusion models prone to off-manifold latents that lead to inaccurate object structures; and (2) the encoder’s inherently weak pixel-level reconstruction hinders the generator from learning accurate fine-grained geometry and texture. In this paper, we propose a systematic framework to adapt understanding-oriented encoder features for generative tasks. 
We introduce a semantic–pixel reconstruction objective to regularize the latent space, enabling the compression of both semantic information and fine-grained details into a highly compact representation (96 channels with $16\times16$ spatial downsampling). This design ensures that the latent space remains semantically rich and achieves state-of-the-art image reconstruction, while remaining compact enough for accurate generation.
Leveraging this representation, we design a unified Text-to-Image (T2I) and image editing model. Benchmarking against various feature spaces, we demonstrate that our approach achieves state-of-the-art reconstruction, faster convergence, and substantial performance gains in both T2I and editing tasks, validating that representation encoders can be effectively adapted into robust generative components.

\end{abstract}

%% file: sec/1_intro.tex
\section{Introduction} \label{sec:intro}

Representation encoders trained via self-supervision~\citep{dino,dinov2,dinov3,moco,mae}  or contrastive learning~\citep{clip,siglipv2,perception-encoder} have established themselves as the cornerstone of visual understanding. They produce highly discriminative, semantic-rich features that generalize exceptionally well, enabling efficient adaptation to downstream tasks with limited data~\citep{liu2023visual,liu2024llavanext}. From dense prediction tasks to complex reasoning in Large Vision Language Models, these encoders have become the universal bedrock of visual analysis. Yet, despite this pervasive dominance, these powerful representations have yet to conquer the generative domain.

Instead, state-of-the-art generative systems predominantly rely on Variational Autoencoder (VAE)~\citep{kingma2013auto}, which operates on low-level, compact latents trained with a  pixel reconstruction objective. These VAE latents lack the high-level semantic structure of representation encoders, forcing diffusion models to learn visual concepts from scratch and necessitating massive computational resources~\citep{esser2024scaling}. To bridge this divide and achieve the long-sought goal of unifying perception and generation, a natural question arises:
\emph{can we migrate the generative modeling space from VAE latents to the representation-encoder space, enabling diffusion models to directly benefit from the representation encoder’s discriminative and semantically structured features?}

\begin{wrapfigure}{R}{0.5\textwidth}
\vspace{-7mm}
\begin{center}
\includegraphics[width=1\linewidth]{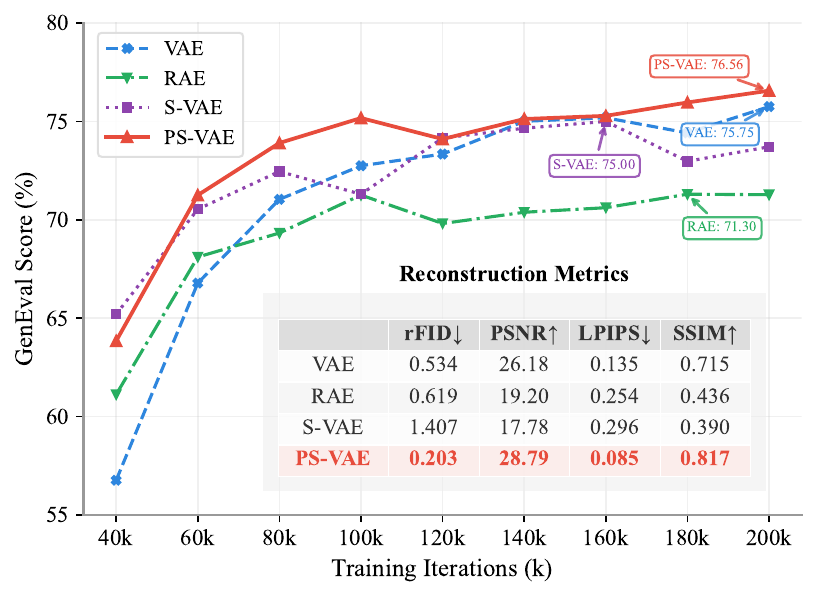}
\end{center}
\vspace{-4mm}
\caption{\textbf{Reconstruction and generation performance across different generation spaces.}
Compared to vanilla \textcolor{Blue}{\texttt{VAE}}, \textcolor{ForestGreen!60}{\texttt{RAE}} improves generation coverage speed but quickly saturates due to its unconstrained semantic space and weak reconstruction.
To address this, we project RAE features into a compact 96-channel latent space with a semantic reconstruction objective, forming \textcolor{Purple!50}{\texttt{S-VAE}}, which mitigates off-manifold issues and improves generation performance.
Finally, \textcolor{Red}{\ours} further augments the semantic latent space with pixel-level reconstruction, enriching structural and texture details and achieving superior performance in both reconstruction and generation.}\label{fig:teaser}
\vspace{-6mm}
\end{wrapfigure}

Recent work, RAE~\citep{rae}, offers a pioneering answer to this question. By redesigning the DiT architecture to handle high-dimensional features, it successfully enables generation within the representation space, achieving impressive results on the class-conditional ImageNet benchmark~\citep{russakovsky2015imagenet}. However, the efficacy of this paradigm does not easily translate to open-world applications. When extended to practical text-to-image synthesis and complex instruction-based editing tasks, RAE exhibits significant performance limitations compared to mature VAE-based baselines, as highlighted in~\Cref{fig:teaser}.

To uncover the root causes of this degradation, we analyze the behavior of the representation space both experimentally and theoretically in~\Cref{ana:rae}, identifying two key issues: insufficient compact regularization of representation features, leading to off-manifold latent generation, together with weak pixel-level reconstruction, which prevents the generator from learning accurate geometry and texture.

The first issue is that generation over representation features is performed in an unconstrained space without compact regularization, leading to a mismatch between the high dimensionality of representation features and their much lower intrinsic information content. Training on such a redundant high-dimensional space makes the diffusion model prone to producing off-manifold latents\footnote{We define ``off-manifold'' latents as features falling into undefined/OOD regions where image decoding becomes unreliable.}, ultimately leading to inaccurate and structurally distorted objects. We verify this phenomenon by visualizing off-manifold outliers through a toy fitting experiment and a theoretical analysis (\Cref{ana:rae} and \Cref{fig:rae_ps_ana}). This observation motivates us to regularize the generative space: we propose  \texttt{S-VAE}, which maps the frozen representation features into a compact, KL-regularized latent space~\citep{rombach2022high} via a semantic autoencoder. This constraint effectively eliminates off-manifold outliers, ensuring that generated latents remain within the valid decoding manifold and thereby improving generation performance (as shown by the \textcolor{Purple!50}{purple} line in~\Cref{fig:teaser}).

The second issue arises from the training objective of representation encoders, which focuses on producing sufficiently discriminative features for understanding rather than preserving detailed structure and fine-grained visual information required for generation. Consequently, even within the regularized \texttt{S-VAE} space, the model struggles to synthesize realistic fine-grained textures and precise small-scale structures. 
To address this, we unfreeze the encoder and jointly optimize it with a pixel-level reconstruction loss on the input image and a semantic reconstruction loss defined on the outputs of the original frozen pretrained encoder. This encourages the encoder to maintain fine-grained details during the computation of strong semantic representations, yielding our final Pixel–Semantic VAE (\ours\Cref{fig:teaser}).

Specifically, we instantiate \ours with a 96-channel latent design based on DINOv2~\citep{dinov2}.
Compared to vanilla VAEs such as MAR-VAE~\cite{li2024autoregressive}, this architecture achieves state-of-the-art reconstruction quality, significantly improving rFID (0.534 $\to$ 0.203), PSNR (26.18 $\to$ 28.79), and SSIM (0.715 $\to$ 0.817).
It also outperforms MAR-VAE in text-to-image generation, exhibiting faster convergence and superior final performance (GenEval~\citep{geneval}: 75.8 $\to$ 76.6; DPG-Bench~\citep{dpg}: 83.2 $\to$ 83.6).
Most notably, on the challenging instruction-based image editing task—requiring both accurate image understanding and faithful instruction execution—\ours delivers a substantial improvement, boosting the editing reward from 0.06 to 0.22. We also validate our method on SigLIP2~\cite{siglipv2}, which is used in Bagel~\cite{deng2025bagel}, observing consistent generation behavior. Importantly, the fine-tuned encoder retains strong understanding ability without any LLM fine-tuning, highlighting the potential of our approach for unifying the encoder for both understanding and generation.

In summary, our contributions are:
\begin{itemize}[leftmargin=*]

\item \textbf{Conceptual insight.} We are the first to show that standard representation encoders are \emph{not} directly suitable for text-to-image generation or instruction-based editing. Through comprehensive analysis, we identify two fundamental issues: off-manifold generation arising from unconstrained feature spaces and poor pixel reconstruction fidelity inherent to discriminative pre-training.

\item \textbf{Methodological design.} We propose a principled approach that transforms the original unconstrained feature space into a compact generative latent space using a semantic autoencoder, and further enriches fine-grained structural and textural details through a pixel-reconstruction objective. The resulting 96-channel latent space, built on DINOv2-B, achieves state-of-the-art performance in both reconstruction and generation, and generalizes effectively to SigLIP2, supporting its potential as a unified encoder for understanding and generation.

\item \textbf{Unified evaluation.} We establish a standardized evaluation pipeline for fair and controlled comparison across different generation space designs on text-to-image and image-editing tasks, and empirically demonstrate the superior performance of \ours under this framework.

\end{itemize}

%% file: sec/2_related_work.tex
\vspace{-3mm}
\section{Related work}
\label{sec:formatting}
\vspace{-2mm}
\noindent  \textbf{Representation Encoders for Visual Understanding} 
Representation learning is foundational to modern visual understanding. By mapping raw image data into a discriminative feature space, pre-trained encoders (such as DINO~\cite{dino, dinov2, dinov3}, SigLIP~\cite{siglip, siglipv2}, and Perception Encoder~\cite{perception-encoder}) enable a wide array of downstream tasks, including classification, object detection, segmentation, and Vision-Language multi-modality modeling~\cite{liu2023visual, Cambrian-1}.  These powerful encoders are typically obtained through two major paradigms: self-supervised learning~\cite{simclr, moco, byol, dino, dinov2, dinov3} and image–text contrastive learning~\cite{clip, siglip, perception-encoder}. Critically, since these representations are optimized mainly for discrimination, they effectively act as lossy compressors, discarding high-frequency pixel details that are non-essential for semantic understanding but vital for accurate synthesis. Furthermore, the resulting feature space is typically high-dimensional and unconstrained (non-regularized). This lack of generative regularization makes them susceptible to the off-manifold generation issue when used directly as a diffusion target (as shown by our analysis in~\Cref{sec:intro}), which prevents their straightforward adaptation to generative modeling tasks.

\noindent \textbf{VAEs for Visual Generation} Variational Autoencoders (VAEs)~\citep{kingma2013auto} are fundamental components of Latent Diffusion Models (LDMs)~\citep{rombach2022high}, primarily serving to reduce the computational cost of high-resolution generation. However, VAEs are trained mainly on a pixel-reconstruction objective, which often yields a latent space focused predominantly on low-level structural details rather than high-level semantic concepts. This forces the subsequent diffusion model to learn complex visual concepts from scratch, necessitating massive computational resources. 
While early LDM work evaluated VAEs almost exclusively via reconstruction fidelity~\cite{rombach2022high}, more recent studies have recognized that the topological properties of the latent space are crucial for robust generation. This realization has prompted efforts to introduce explicit regularization~\cite{kouzelis2025eq, skorokhodov2025improving, yao2025vavae}, often via KL-divergence constraints, to encourage better latent utilization and stability. Our work extends this idea by explicitly designing a compact, KL-regularized latent space that is directly informed by high-level semantics, thus combining the generative stability of VAEs with the discriminative power of foundation models.

\noindent \textbf{Unifying Feature Spaces for Generation and Understanding}
Recent efforts to bridge the gap between discriminative and generative feature spaces generally follow two distinct strategies. The first focuses on aligning standard VAE latents with representation encoders through representation-alignment objectives, treating the encoder as a soft semantic constraint (e.g., \cite{yao2025vavae, xu2025exploring}). The second strategy, which is more closely related to this work, aims to construct a generative feature space directly from representation encoders~\cite{chen2025aligning, lu2025atoken, yue2025uniflow, shi2025latent, rae}. ~\cite{chen2025aligning,lu2025atoken,yue2025uniflow} do not explicitly require the latent space to fully preserve the original semantic structure of the representations.
More closely related to our study, SVG~\cite{shi2025latent} and RAE~\cite{rae} propose diffusing directly over raw, unconstrained, high-dimensional semantic features. While this approach captures powerful semantics, we identify two critical failures preventing its widespread application: insufficient compact regularization of representation features, leading to off-manifold latent generation, together with weak pixel-level reconstruction, which prevents the generator from learning accurate geometry and texture. Thus, we propose \ours, which introduces a principled intermediate step: creating a compact, KL-regularized semantic latent space and then applying a joint pixel-reconstruction objective to enrich it with high-fidelity  details. This design achieves superior performance in tasks requiring both  semantic understanding and precise structural control, \eg instruction-based image editing.  Related work in the autoregressive paradigm~\cite{ma2025unitok, song2025dualtoken, lin2025toklip, han2025tar} is discussed in the \supp, as it follows a fundamentally different modeling approach.

%% file: sec/3_method.tex
\section{Method} 
\label{exp:method}

\subsection{Overview}
In this section, we begin by providing a detailed analysis of the reconstruction and generation behavior of RAE~\citep{rae}. We theoretically and experimentally validate that its extremely high-dimensional feature space makes the diffusion model prone to generating off-manifold latent features that lie outside the training support of the pixel decoder. Furthermore, RAE's inherently poor reconstruction quality hinders the generative model from learning accurate object structures and fine-grained textures, making it unsuitable for high-fidelity tasks such as instruction-based editing.
To address these fundamental challenges, we subsequently introduce our step-wise strategy for preparing the representation encoder for generation by mapping both pixels and representations into a unified, compact latent space.
Finally, we present a Deep-Fusion architecture for text-to-image generation and image editing, establishing a fair benchmarking framework for different generative latent spaces. Unless otherwise specified, we follow the settings of RAE~\cite{rae} and use a DINOv2-B~\cite{dinov2} encoder for feature extraction.

\subsection{Analysis of RAE} \label{sec:rae_analysis}

\begin{figure*}[t]
\begin{center}
    \includegraphics[width=\textwidth]{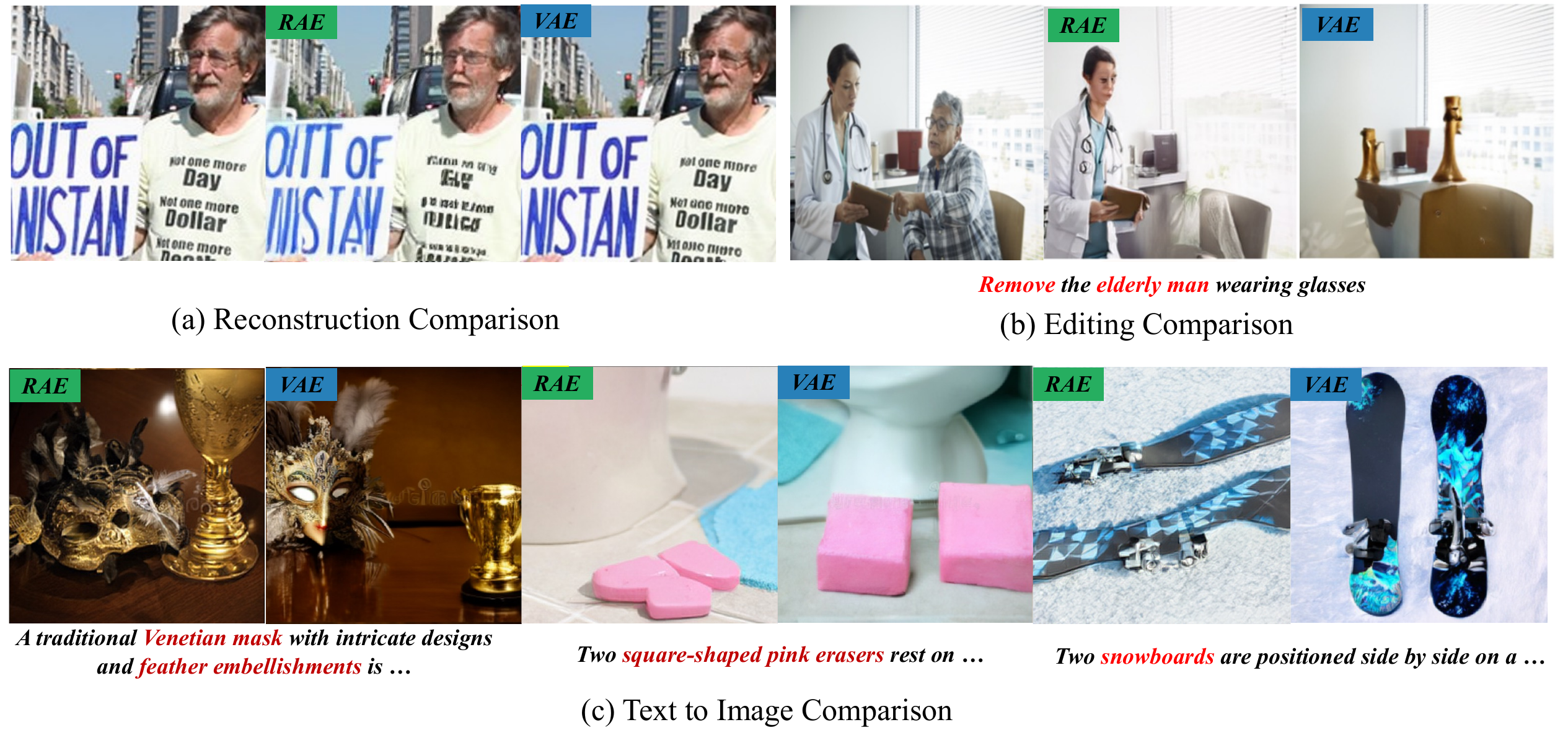}
\end{center}
\caption{\textbf{Visualization comparison between RAE and VAE.}
(a) RAE shows a noticeable gap in reconstruction performance compared to VAE.
Benefiting from its rich semantic representation, RAE demonstrates stronger prompt-following ability in image editing tasks that require understanding the input image (b). However, its poor reconstruction quality limits practical usability, as it fails to preserve fine-grained and consistent details from the input image. Counterintuitively, in text-to-image generation, RAE exhibits severe structural and texture artifacts and substantially lags behind VAE (c), with a performance gap far larger than that observed in reconstruction.}\label{fig:rae_ana}
\vspace{-3mm}
\end{figure*}

\noindent\textbf{Comparison: RAE \vs VAE}\label{ana:rae}
To analyze the reconstruction and generation behavior of RAE~\cite{rae}, we conduct a series of visualization and benchmarking experiments, comparing it with a vanilla VAE~\cite{li2024autoregressive} using the same $16\times16$ spatial compression. 
We first compare the reconstruction performance. As shown in~\Cref{fig:rae_ana}(a), RAE exhibits a notable shortfall in reconstruction quality compared to VAE, often introducing artifacts in regions such as faces and text. This aligns with the quantitative findings in~\Cref{tab:rae_to_our}: although RAE achieves a comparable rFID, its SSIM and PSNR are significantly lower. 
This behavior is expected, as the DINOv2 encoder used in RAE is trained with a purely discriminative objective and does not explicitly optimize for reconstruction. 
To further validate their impact on generation, we conduct both text-to-image generation and image editing tasks (All equipped with a wide DDT head as in RAE~\cite{wang2025ddt,rae}). Notably, in text-to-image generation, despite faster coverage (as shown in ~\Cref{fig:teaser}) enabled by its strong semantic feature space, RAE still suffers from severe structural and texture artifacts (see~\Cref{fig:rae_ana}(c)) and substantially underperforms VAE, resulting in a much poorer performance on benchmarks such as GenEval~\cite{geneval}. For image editing tasks, RAE demonstrates a superior capacity for prompt-following in edits that necessitate semantic comprehension of the input image (see~\Cref{fig:rae_ana}(b)), but its poor reconstruction quality limits detail preservation.

In conclusion, we empirically observed that RAE exhibits inferior reconstruction quality compared to VAE, limiting its ability to preserve fine-grained details in both image generation and editing. Conversely, its semantically rich latent space facilitates faster coverage in text-to-image generation and superior prompt-following in editing tasks. While these results largely align with expectations, we noted one counterintuitive finding: \emph{ objects generated based on RAE space suffer from severe artifacts compared to VAE, the severity of which far exceeds what would be predicted by the reconstruction gap alone.}

\begin{figure}[!t]
\begin{center}
\includegraphics[width=1\linewidth]{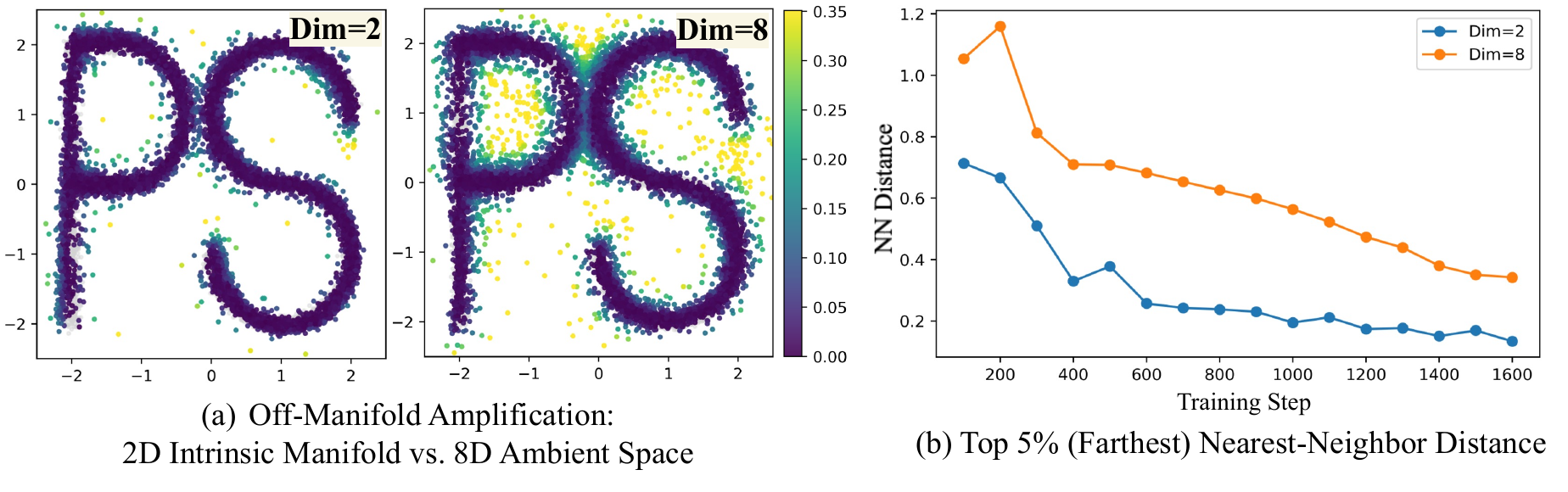}
\end{center}
\vspace{-3mm}
\caption{\textbf{Off-manifold behavior varies significantly with feature dimensionality.}
We construct a 2D `PS'-shaped distribution and embed it into an 8D ambient space, yielding two learning settings with intrinsic dimension 2 and ambient dimension 8.
\textbf{(a)} The 8D setting produces substantially more off-manifold samples than the intrinsic 2D space.
\textbf{(b)} We measure the mean nearest-neighbor distance of the top 5\% tail samples and observe that samples generated in 8D deviate much farther from the data manifold, indicating stronger off-manifold drift.}
\label{fig:rae_ps_ana}
\vspace{-4mm}
\end{figure}

\noindent\textbf{Analysis of Off-Manifold Behavior in RAE}
The structural and texture artifacts in RAE far exceed typical reconstruction errors, suggesting a fundamental problem in the generation process. We hypothesize that these artifacts arise because the diffusion model, trained on the high-dimensional RAE feature space, generates off-manifold samples. These samples reside outside the training distribution of the pixel decoder, leading to the sub-optimal decoded results. We trace the root cause to the discrepancy between the high dimensionality of DINOv2 features and their lower intrinsic information content.

To rigorously analyze the difficulty of learning a low-dimensional intrinsic manifold embedded in a high-dimensional space, we model the generative dynamics of an $h$-dimensional feature space containing an $l$-dimensional manifold ($h>l$). Let $z \in \mathbb{R}^l$ denote the latent data and $x = Qz \in \mathbb{R}^h$ denote the observed data, where $Q \in \mathbb{R}^{h\times l}$ is a column-orthonormal mapping ($Q^\top Q = I_l$). The forward diffusion processes for $x$ and $z$ are coupled: $x_t = (1-t) x_0 + t \epsilon_h$ implies that the projected variable $z_t = Q^\top x_t$ follows $z_t = (1-t) z_0 + t \epsilon_l$, where $\epsilon_l = Q^\top \epsilon_h$. Beyond the coupling of the forward processes, the optimal denoising objectives are strictly related. We denote the optimal velocity estimators for the intrinsic and embedded processes as $v_{z, \theta}$ and $v_{x, \theta}$, respectively. These are defined as the expected velocity targets given the noisy states:
\begin{equation*}
v_{z, \theta} (z_t) = \mathbb{E}[\epsilon_l-z_0 | z_t],\quad v_{x, \theta} (x_t) = \mathbb{E}[\epsilon_h-x_0 | x_t]
\end{equation*}
By projecting the signal in the embedded high dimensions onto the data manifold and its orthogonal complement, we can express the high-dimensional estimator $v_{x, \theta}$ purely in terms of the low-dimensional estimator $v_{z, \theta}$ plus a residual term:
\begin{equation}
\label{eq:decomposition}
v_{x, \theta} (x_t) = Q v_{z, \theta} (Q^\top x_t) + \frac{1}{t} (I - QQ^\top)x_t
\end{equation}

Equation~\ref{eq:decomposition} reveals a fundamental disparity in learning difficulty.
The first term represents the generative flow along the intrinsic manifold. While this component is intrinsically low-dimensional, the model operating in the ambient space must implicitly learn the projection ($Q^\top$) and embedding ($Q$) operations to resolve it. This imposes a significant burden of manifold discovery—identifying the sparse data subspace within the vast high-dimensional ambient space—a challenge that is entirely bypassed when diffusing directly in the intrinsic latent coordinates.

The second term consists purely of Gaussian noise in the orthogonal subspace, which forces the network to learn an identity-like mapping that conveys no semantic information, resulting in inefficient use of model capacity. This also helps explain the behavior observed in RAE~\cite{rae}: when the model dimensionality is smaller than the input dimensionality, the network struggles to fit even a single example. \emph{High-dimensional Gaussian noise is full-rank and cannot be compressed without loss, which forces the model to allocate sufficient capacity to transmit noise.}  As a result, an intrinsic information bottleneck emerges when the model width is smaller than the ambient feature dimension.\label{sec:wide_head} This also explains why the wide DDT-Head design~\cite{wang2025ddt, rae}, which incorporates a long skip connection from the input x, substantially improves the performance of RAE.

To validate this theoretical analysis, we investigate diffusion training in a high-dimensional feature space ($\text{h}=8$) that implicitly contains a lower-dimensional intrinsic manifold ($\text{l}=2$). We construct a ground-truth 2D ``PS''-shaped distribution $\boldsymbol{z}$ and embed it into $\mathbb{R}^8$ via a linear isometric mapping $\boldsymbol{x} = Q\boldsymbol{z}$, where $Q \in \mathbb{R}^{8\times 2}$ has orthonormal columns. In this setup, $Q$ acts as the linear decoder defining the manifold. We then train identical 256-channel MLP-based diffusion models separately on the intrinsic 2D data and the embedded 8D data. For evaluation, we project the generated 8D samples back to the 2D plane using the linear encoder $Q^\top$ (noting that $Q^\top Q = I_2$, which perfectly recovers on-manifold data). 
As shown in~\Cref{fig:rae_ps_ana}, learning in the 8D ambient space results in slower convergence and a degradation in sample quality. Nearest-neighbor distance evaluation against the ground truth reveals that the top 5\% tail samples from the 8D model deviate significantly from the true manifold. Despite sharing the same intrinsic geometry, the unconstrained high-dimensional representation amplifies off-manifold behavior. This confirms that discovering and training on the intrinsic low-dimensional isomorphic distribution is essential for stabilizing diffusion training and eliminating generation artifacts.

\subsection{Make Representation Encoders Ready}

Building on our analysis of RAE~\cite{rae}, we first address the off-manifold problem, identified as the primary limitation. Subsequently, we enhance reconstruction fidelity to improve the text-to-image performance and enable detail-sensitive applications such as image editing.

\begin{figure*}[t]
\begin{center}
    \includegraphics[width=1\textwidth]{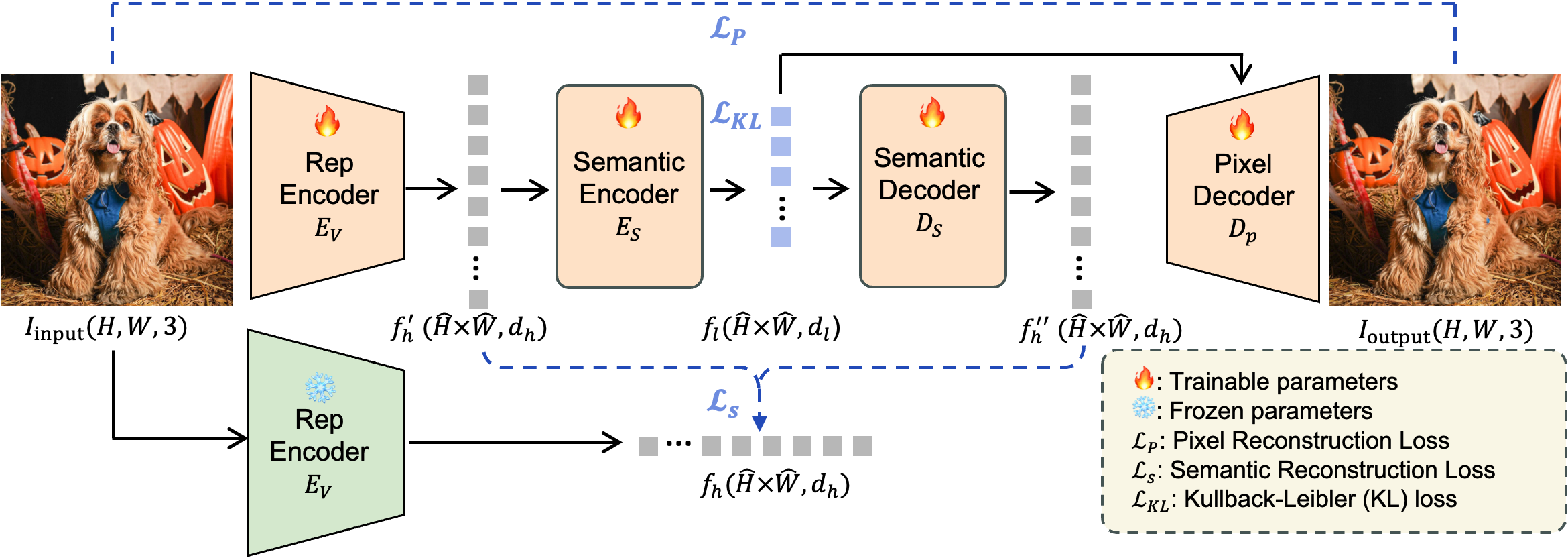}
\end{center}
\caption{\textbf{Compact latent space construction for preserving semantic structure and fine-grained details} We first regularize the unconstrained representation-encoder feature space by freezing the encoder and training a semantic VAE using only the $\mathcal{L}_s$ and $\mathcal{L}_\mathrm{kl}$; during this stage, the pixel decoder is trained on the detached semantic latent with pixel reconstruction loss $\mathcal{L}_\mathrm{P}$. After semantic reconstruction converges, we unfreeze all components and allow the pixel decoder to backpropagate the gradient into the encoder, ensuring that the representation encoder captures fine-grained details of the input image.
}\label{fig:vae_arch}
\vspace{-3mm}
\end{figure*}

The overall training framework of \ours is illustrated in \Cref{fig:vae_arch}. Given an input image $I_{\mathrm{input}} \in \mathbb{R}^{H \times W \times 3}$, we first extract a semantic feature map $f'_h \in \mathbb{R}^{\hat{H} \times \hat{W} \times d_h}$ using a pretrained Representation Encoder, \eg DINOv2-B~\citep{dinov2}. As discussed in \Cref{sec:rae_analysis}, $f'_h$ is a high-dimensional, unconstrained representation, where its dimension $d_h = 768$. 

\begin{wrapfigure}{r}{0.4\textwidth}
\begin{center}
\vspace{-4mm}
\includegraphics[width=\linewidth]{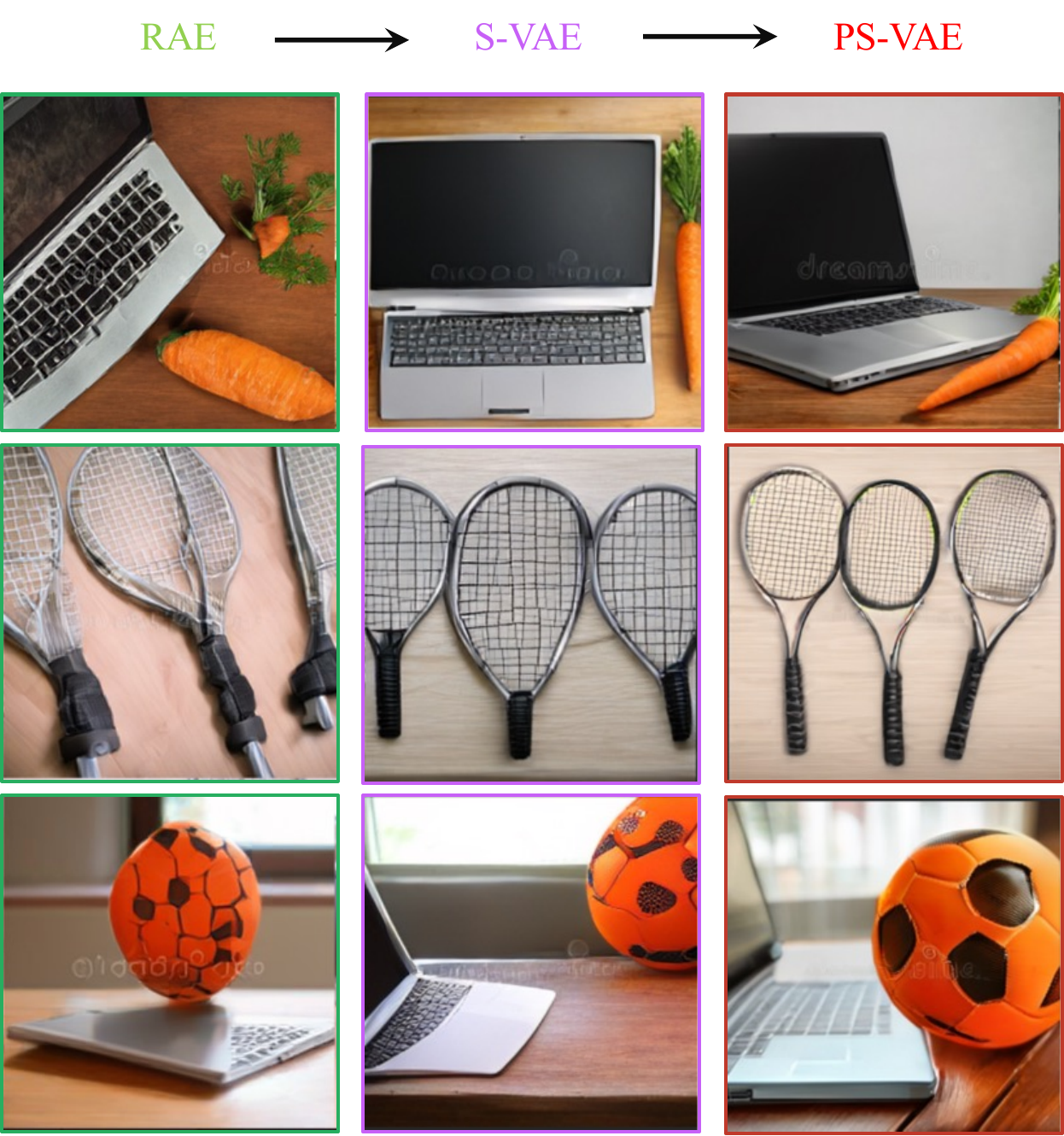}
\end{center}
\vspace{-2mm}
\caption{\textbf{Visual comparison of generated examples across progressively improved latent spaces (RAE $\rightarrow$ S-VAE $\rightarrow$ PS-VAE).}
Artifacts are gradually reduced, with step-by-step improvements in texture and structure.}\label{fig:rae_to_ours}
\vspace{-4mm}
\end{wrapfigure}

To address the off-manifold problem, we introduce a semantic VAE~(\textcolor{purple!50}{S-VAE}) that maps the high-dimensional unconstrained feature space $f'_h$ to a compact latent space $f_l \in \mathbb{R}^{\hat{H} \times \hat{W} \times d_l}$ via an encoder $E_s$. Here, $d_l=96$ ($d_l \ll d_h$). Then, another semantic decoder $D_s$ is adopted to reconstruct the latent back to the original feature $f''_h$.
Both the semantic encoder $E_s$ and decoder $D_s$ are optimized with a semantic reconstruction loss $\mathcal{L}_s$, which combines an $\ell_2$ loss and a cosine similarity loss on features, while the latent is further regularized by a Kullback--Leibler divergence loss $\mathcal{L}_{\mathrm{KL}}$ following~\cite{rombach2022high}.
The encoder and decoder share a symmetric design with three Transformer blocks inherited from the representation encoder and an MLP projection layer for dimensionality adjustment.
 For the evaluation of \textcolor{purple!50}{S-VAE}, we additionally train a pixel decoder that reconstructs the output image $I_{\mathrm{output}}$ from the detached semantic latent $f_l.\mathrm{detach()}$ via the pixel reconstruction loss $\mathcal{L}_P$(The pixel reconstruction loss $\mathcal{L}_P$ follows~\cite{rombach2022high}.). A diffusion model is further trained on this semantic VAE (\textcolor{purple!50}{S-VAE}) latent space.
As shown in~\Cref{fig:rae_to_ours} and~\Cref{tab:rae_to_our}, both visual quality and quantitative results are substantially improved, despite a slight performance drop in reconstruction fidelity.
This result confirms that the primary limitation lies in the off-manifold issue rather than reconstruction quality.

To enhance image reconstruction without compromising the semantic structure of the latent space, we unfreeze the representation encoder during pixel decoder training. By removing the \texttt{detach} operation in $f_l$, we enable gradients to propagate from the pixel decoder back to the encoder. 
To preserve the pretrained semantic representations during this optimization, we enforce a semantic reconstruction loss on $f'_h$ and $f''_h$ relative to the original encoder, while retaining the KL loss and pixel-reconstruction loss. 
After this training stage, we obtain our Pixel-Semantic VAE (\textcolor{red}{PS-VAE}).

As demonstrated in \Cref{fig:rae_to_ours} and \Cref{tab:rae_to_our}, this strategy significantly improves the reconstruction quality of the representation encoder while preserving its semantic structure. This enables the generation model to learn fine-grained geometry and texture, while the well-preserved semantics ensure fast coverage of text-to-image pretraining and strong instruction-following ability for image editing(as shown in \Cref{fig:all_vae_geneval_dpg}). As a result, both visual quality and quantitative performance are consistently enhanced for text-to-image generation and editing.

\subsection{Generation Architecture}

\begin{wrapfigure}{r}{0.5\textwidth}
\begin{center}
\vspace{-8mm}
\includegraphics[width=1\linewidth]{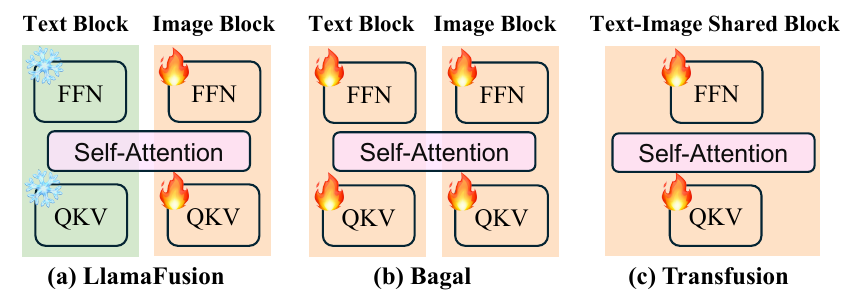}
\end{center}
\vspace{-3mm}
\caption{Block comparison of three deep-fusion architectures.}\label{fig:gen_arch}
\vspace{-3mm}
\end{wrapfigure}

Unified models for generation and understanding are being actively explored. Owing to its strong semantic representation and high-fidelity reconstruction, \ours has strong potential to serve as a unified encoder in such frameworks. For these reasons, we adopt a deep-fusion architecture as our generation paradigm. To investigate which deep-fusion architecture yields superior performance, we first conduct a preliminary ablation study on three popular deep-fusion designs for generation, as illustrated in \Cref{fig:gen_arch}. (a) LlamaFusion~\cite{shi2024lmfusion}, which freezes all language blocks and adds parallel image blocks with identical architecture; (b) Bagel-style models~\cite{deng2025bagel}, which unfreeze both text and image branches to improve multimodal alignment; and (c) Transfusion~\cite{zhou2025transfusion}, which processes image and text tokens jointly using fully shared transformer blocks.

\begin{wraptable}{r}{0.5\columnwidth}
    \vspace{-6mm}
    \centering
    \caption{GenEval scores of Deep-Fusion architectures.}
    \label{tab:deep-fusion}
    \resizebox{0.5\columnwidth}{!}{
    \begin{tabular}{lcc}
        \toprule
        \textbf{Model} & \textbf{Params (M)} & \textbf{GenEval} $\uparrow$ \\
        \midrule
        LlamaFusion             & 857 & 0.524 \\
        Bagel                   & 857 & 0.763 \\
        Transfusion             & 500 & 0.752 \\
        \midrule
        Transfusion + Wide DDT Head & 653 & \textbf{0.762} \\
        \bottomrule
    \end{tabular}}
    \vspace{-3mm}
\end{wraptable}

We  evaluate the three deep-fusion architectures using VAVAE~\cite{li2024autoregressive}
(32-channel latent, stride-16, patch size 1). All fusion blocks are initialized from Qwen2.5-0.5B~\cite{qwen}. We apply 2D positional encoding to the VAE features and inject timestep embeddings into their initial hidden states before feeding them into the LLM backbone, following Bagel~\cite{deng2025bagel}. For text-to-image, we concatenate text embeddings with the noisy image latent. Text tokens use a causal mask, while noisy image latent uses full attention mask. For image editing, we concatenate the clean latents of input images, the instruction text embeddings, and the noisy latents. We apply a full attention mask to the clean and noisy latents, while employing a causal mask for the instruction texts.

Results in Table.~\ref{tab:deep-fusion} indicate that LlamaFusion exhibits a clear bottleneck, likely due to its frozen language branch being unable to adapt to text-to-image generation. Compared to the Transfusion-style block design, the Bagel-style design improves performance by 1.1 but increases parameters by 71\%. Since we only evaluate text-to-image performance across different feature spaces and do not consider preserving language modeling capability, we adopt the TransFusion-style block as our core fusion architecture for better parameter efficiency.

With the fusion block fixed, we incorporate the wide DDT head~\cite{wang2025ddt} from RAE~\cite{rae}, which enhances generation quality in high-channel feature spaces(as we analyzed in \Cref{sec:wide_head}). We validate its effectiveness through consistent gains across multiple VAEs. As shown in Table~\ref{tab:deep-fusion}, the head improves VAVAE (32-channel, stride-16, patch size 1) from 75.2 to 76.2. We observe similar improvements for Flux-VAE~\cite{flux2024} (16-channel, stride-8, patch size 2), which increases from 63.7 to 68.04, and MAR-VAE (16-channel, stride-16, patch size 1), which rises from 72.6 to 75.75. Given these consistent results, we adopt the wide DDT head as a standard component.

%% file: sec/4_exp.tex
\section{Experiments}

In this subsection, we outline our training and inference pipelines, followed by the evaluation protocols for reconstruction, text-to-image generation, and instruction-based image editing. We then present performance results across varying feature spaces to demonstrate the effectiveness of \ours. Subsequently, we analyze the scaling behavior of our 96- and 32-channel variants, showing that larger generation models effectively leverage the rich semantic and pixel-level details preserved in high-channel latent spaces. Finally, we extend our framework by replacing the DINOv2~\cite{dinov2} encoder with SigLIP2~\cite{siglipv2}, highlighting the latter's potential as a unified encoder for both visual understanding and generative modeling.

\subsection{Training and Evaluation Details}

\noindent \textbf{Reconstruction.}
To ensure a fair comparison with prior work, we train our reconstruction models exclusively on ImageNet-1K~\cite{russakovsky2015imagenet}, though we note that future work could benefit from larger, more diverse datasets. Input images are resized and center-cropped to $224 \times 224$. Using a patch size of 14 results in a sequence length of $16 \times 16$, making the computation—in terms of both FLOPs and runtime—significantly more efficient than the VAE-style encoders used in Latent Diffusion Models (LDM)~\cite{rombach2022high}.
Our pixel decoder adopts the LDM architecture~\cite{rombach2022high} and reconstructs images at a resolution of $256 \times 256$. We evaluate performance using rFID, SSIM, PSNR, and LPIPS on the ImageNet-1K validation set. Models are trained with a batch size of 96 and a learning rate of $10^{-4}$.
We employ a two-stage training strategy: first, we freeze the foundation model and train only the semantic encoder and decoder, with the pixel decoder trained on detached semantic latents to prevent interference with semantic compression. In the second stage, we unfreeze all components, allowing gradients from the pixel decoder to backpropagate to both the foundation model and the semantic encoder. The loss weights for $L_s$ and $L_p$ are set to 1 and 0.1, respectively. Further details are provided in the \supp.

\noindent \textbf{Text-to-Image Generation.}
We utilize CC12M-LLaVA-NeXT~\cite{cc12m, cc12m-llavanext} for training, which comprises 10.9 million images with detailed long-form captions~\cite{liu2024llavanext}. Images are resized and center-cropped to $256 \times 256$. We evaluate performance using GenEval~\cite{geneval} and DPG-Bench~\cite{dpg}. GenEval relies on object detection, making it highly sensitive to structure and texture—factors closely tied to human perceptual preference.
If generated objects exhibit geometric inaccuracies or distorted textures, the detector may fail to classify them correctly or produce duplicate detections. As a result, scores can be lower even when the text–image semantic alignment appears correct at a glance.
Conversely, DPG-Bench employs a vision–language model as a judge, prioritizing high-level alignment over fine-grained details. This complementarity allows us to better interpret trade-offs between structural fidelity and semantic alignment.
We train with a batch size of approximately 730, a learning rate of $10^{-4}$, and apply EMA with a decay of 0.9999. Training for 200K iterations ensures convergence across various generative feature spaces. For GenEval, we use the rewritten long-prompt version from Bagel~\cite{deng2025bagel}, consistent with our long-caption training data.

Variations in patch size and channel dimensionality along the sequence length alter the signal-to-noise ratio (SNR) during interpolation between noise and latents. To maintain consistent SNR weighting across feature spaces, we apply a shifted timestep $t' = \frac{\textit{shift\_factor} \cdot t}{1 + (\textit{shift\_factor} - 1) \cdot t}$, where $t$ is sampled from a Logit-Normal distribution~\cite{esser2024scaling,rae}. Since the sequence length is fixed across feature spaces, the shift factor depends only on the latent channel dimension $C_{\text{vae}}$ and patch size $P_{\text{vae}}$: $\textit{shift\_factor} = \sqrt{\frac{C_{\text{vae}} P_{\text{vae}}^{2}}{C_{\text{base}} P_{\text{base}}^{2}}}$, where $C_{\text{base}} = 16$ and $P_{\text{base}} = 1$.
For instance, Flux~\cite{flux2024} ($C=16, P=2$) yields a shift factor of 2, while DINOv2-B ($C=768, P=1$) yields approximately 6.93. 
We conduct ablation studies to confirm that these calculated values are reasonable. For Flux, a factor of 2 yields a better GenEval score compared to factors of 1 and 3, while for RAE, a factor of 6.93 performs best compared to 6 and 8.
We therefore apply this rule to all feature-space and channel-number ablation experiments. During inference, we use 50-step Euler sampling with a timestep shift of 3 and a classifier-free guidance scale of 6.5.

\noindent \textbf{Instruction Editing.}
We utilize the OmniEdit dataset~\cite{wei2024omniedit}, which
contains 1.2 million image–editing pairs spanning seven editing categories: object replacement, object removal, object addition, attribute modification, background swap, environment change, and style transfer. We resume the text-to-image checkpoint as the initialization for the editing training. For evaluation, we adopt the corresponding editing subtasks from GEdit-Bench~\cite{liu2025step1x}  (Background Change, Color Alteration, Material Modification, Style Transfer, Subject Addition, Subject Removal, Subject Replacement, and Tone Transformation), using their provided input images and prompts. Results are evaluated using EditingReward~\cite{wu2025editreward}, a state-of-the-art image editing scoring model that assesses both visual consistency and instruction adherence. This setup ensures reliable and reproducible evaluation. Training and inference settings match the text-to-image configuration, with models trained for 50k iterations and the best-performing checkpoints reported.

\subsection{Reconstruction and Generation Performance of Different Feature Space}

\begin{table*}[!t]
\centering
\caption{\textbf{Comparison of reconstruction and generation performance. }
The best results are shown in \textbf{bold} and the second-best are \underline{underlined}. Flux-VAE (stride 8) is listed for reference, and all other results correspond to the feature space with a stride 16.}\label{tab:all_vae}
\resizebox{\columnwidth}{!}{
\begin{tabular}{l|cccc|ccc}
\toprule
\textbf{Method} 
& \textbf{rFID}~(↓) 
& \textbf{PSNR}~(↑) 
& \textbf{LPIPS}~(↓) 
& \textbf{SSIM}~(↑) 
& \textbf{GenEval}~(↑)  
& \textbf{DPG-Bench}~(↑) 
& \textbf{Editing Reward}~(↑)  \\
\midrule

Flux-VAE~\cite{flux2024}

& \textbf{0.175 }
& \textbf{32.86}
& \textbf{0.044}
& \textbf{0.912}
& 68.04
& 78.98 
& -0.271 \\

\midrule

MAR-VAE~\cite{li2024autoregressive}

& 0.534 
& 26.18 
& 0.135 
& 0.715 
& 75.75
& 83.19 
& 0.056 \\

VAVAE~\cite{yao2025vavae}

& 0.279
& 27.71
& 0.097
& 0.779
& 76.16
& 82.45 
& \underline{0.227} \\

RAE~\cite{rae}

& 0.619 
& 19.20 
& 0.254 
& 0.436 
& 71.27
& 81.72 
& 0.059 \\

\midrule

\textbf{$\ours_{32c}$} 

& 0.584 
& 24.53 
& 0.168 
& 0.662 
& \underline{76.22} 
& \textbf{84.25} 
& \textbf{0.274} \\

\textbf{$\ours_{96c}$} 

& \underline{0.203} 
& \underline{28.79} 
& \underline{0.085} 
& \underline{0.817} 
& \textbf{76.56}
& \underline{83.62} 
& 0.222 \\

\bottomrule
\end{tabular}}
\end{table*}

\begin{figure*}[t]
\centering
\begin{minipage}[b]{0.33\linewidth}
    \centering
    \includegraphics[width=\linewidth]{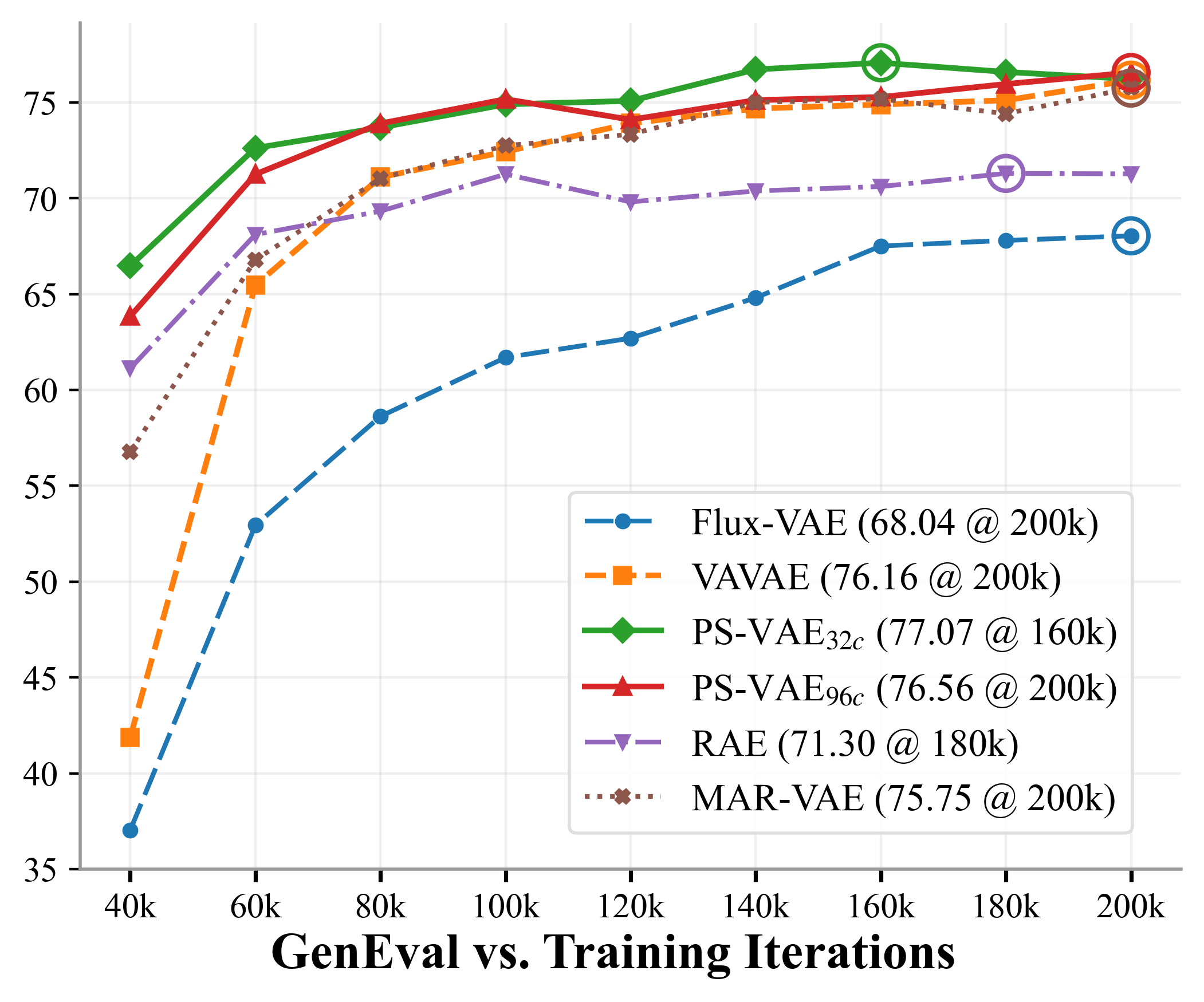}
    \vspace{-2mm}

\end{minipage}
\hfill
\begin{minipage}[b]{0.33\linewidth}
    \centering
    \includegraphics[width=\linewidth]{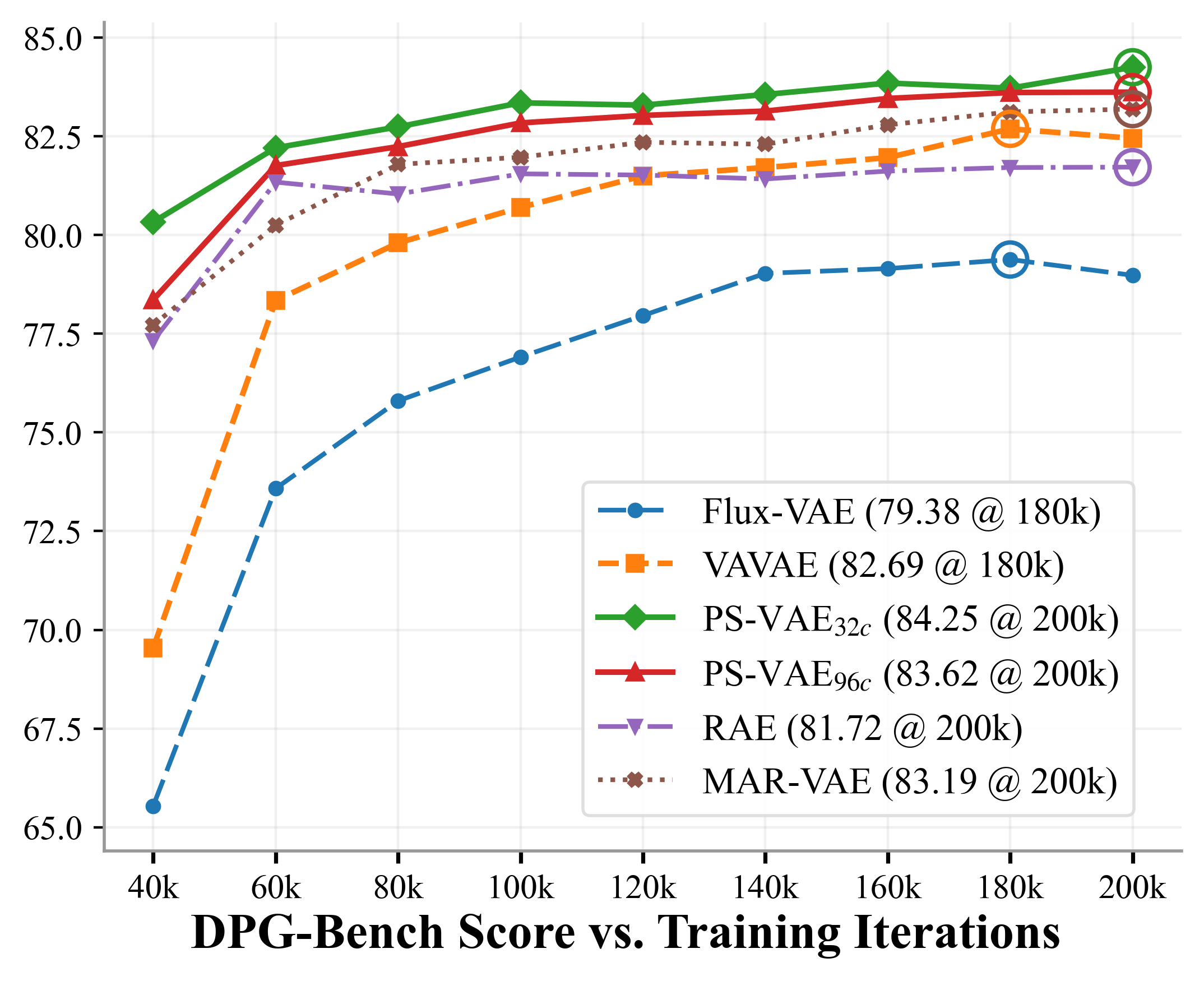}
    \vspace{-2mm}

\end{minipage}
\hfill
\begin{minipage}[b]{0.33\linewidth}
    \centering
    \includegraphics[width=\linewidth]{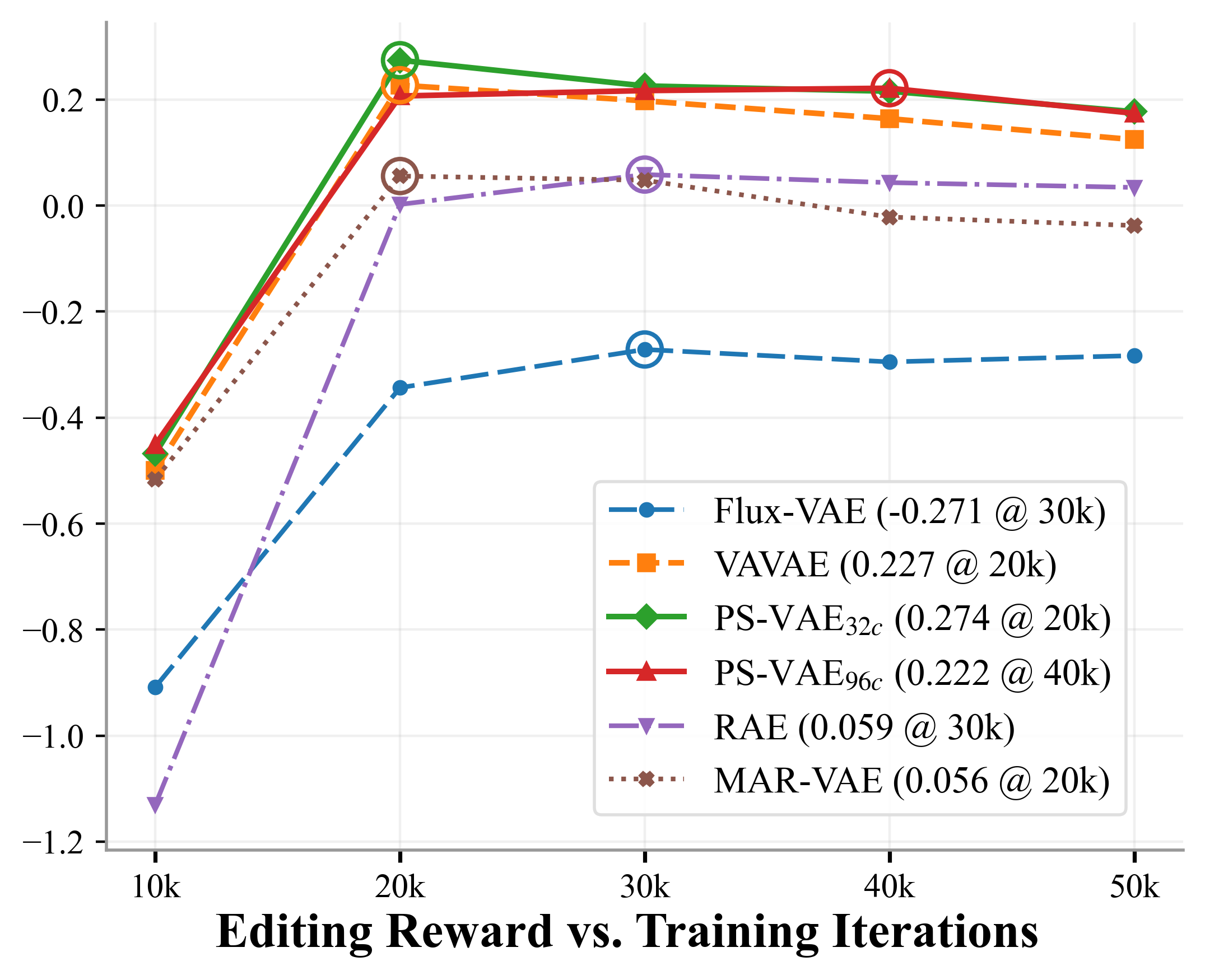}
    \vspace{-2mm}

\end{minipage}

\vspace{-4mm}
\caption{\textbf{Coverage curves for generation and editing tasks across different feature spaces.}
By jointly providing rich semantics and state-of-the-art reconstruction fidelity, \ours consistently outperforms semantic-only RAE and pixel-only VAEs across both generation and editing benchmarks.
The strong semantic structure and well-regularized latent space of \ours enable significantly faster convergence during text-to-image training, while also facilitating better image understanding and, consequently, stronger instruction-following in image editing.
Meanwhile, higher reconstruction fidelity leads to more realistic structures and textures in text-to-image generation, improved detail consistency between the edited and input images during editing, and thus better overall performance in both tasks.
}\label{fig:all_vae_geneval_dpg}
\vspace{-4mm}
\end{figure*}

As shown in \Cref{tab:all_vae}, our 96-channel $\ours_{96c}$ achieves the highest reconstruction quality among all stride-16 VAEs, trailing only Flux-VAE, which benefits from a finer stride of 8. In generation and editing tasks, both $\ours_{32c}$ and $\ours_{96c}$ significantly outperform RAE with the same training budget. Specifically, $\ours_{32c}$ achieves top performance on DPG-Bench~\cite{dpg} and Editing Reward~\cite{wu2025editreward}, ranking second on GenEval~\cite{geneval}. Meanwhile, $\ours_{96c}$  leads on GenEval and ranks second and third on DPG-Bench and Editing Reward, respectively, maintaining a clear advantage over the RAE baseline. Besides, benefiting from a well-constrained and semantically rich latent space, \ours converges faster than RAE and other VAEs (see \Cref{fig:all_vae_geneval_dpg}). Furthermore, enhanced detail fidelity enables \ours to surpass standard VAE performance, highlighting the distinct advantage of the pixel–semantic constrained latent space.

\begin{figure*}[!t]
\begin{center}
\includegraphics[width=0.9\linewidth]{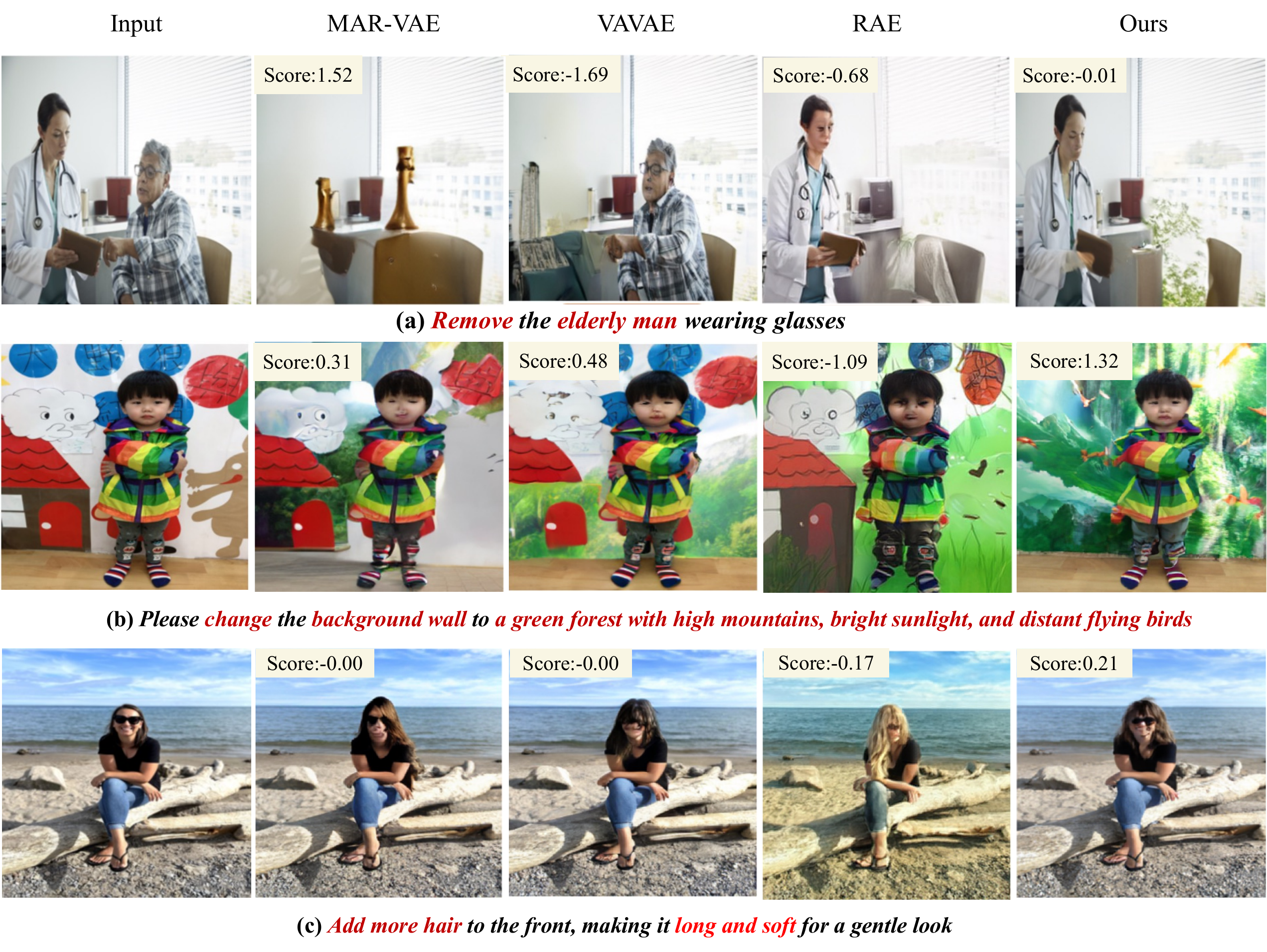}
\end{center}

\caption{\textbf{Editing visual examples of models trained on different feature spaces.} 
As shown in (a), both \ours and RAE exhibit reasonable visual grounding, correctly identifying the elderly man and the background wall in (a). However, RAE’s performance is strongly limited by its weak reconstruction ability, resulting in inconsistent details with the input image (a,b,c). In contrast, benefiting from strong semantic alignment and high-fidelity reconstruction, \ours achieves accurate instruction following while preserving consistent visual details between the input and edited images, such as the human face in (a,b,c).}
\label{fig:edit_samples}
\vspace{-3mm}
\end{figure*}

Meanwhile, we observed in \Cref{tab:all_vae} and \Cref{fig:edit_samples} that VAEs trained solely on pixel reconstruction objectives (e.g., MAR-VAE and Flux-VAE) exhibit significantly lower prompt-following capabilities than models with semantically structured latent spaces, \eg $\ours_{32c}$, $\ours_{96c}$, and VAVAE. We hypothesize that instruction-based editing couples two subtasks: semantic comprehension of the input latent and faithful generation based on the prompt. Consequently, a semantically organized latent space facilitates source image interpretation and improves instruction adherence. This aligns with findings from Bagel~\cite{deng2025bagel}, where the injection of SigLIP2~\cite{siglipv2} features similarly enhances performance. We further verify this observation by the clear drop in editing performance when the semantic reconstruction loss is removed from our training pipeline (denoted as P-VAE), decreasing from 0.22 for PS-VAE to 0.04, as shown in \Cref{tab:rae_to_our}.
Besides, we also observed that RAE’s editing performance is constrained by weak reconstruction capabilities, resulting in visual inconsistencies relative to the input, while \ours effectively balances semantic understanding with fine-grained detail preservation. That's to say, \ours retains the strong instruction-following capability of RAE (see \Cref{fig:edit_samples}.a) while achieving superior consistency in fine-grained regions such as facial features (see \Cref{fig:edit_samples}.a,b,c).

\subsection{Scaling Behavior of Generative Models across PS-VAE Channel Dimensions}

As shown in \Cref{fig:all_vae_geneval_dpg} and \Cref{tab:all_vae}, both $\ours_{32c}$ and $\ours_{96c}$ achieve state-of-the-art generation performance. While $\ours_{96c}$ provides better reconstruction quality, it slightly underperforms $\ours_{32c}$ in generation metrics, likely due to limited model capacity when modeling excessive fine-grained details. We further examine whether increasing model capacity can mitigate this effect.

\begin{figure*}[t]
    \centering
    \begin{minipage}[b]{0.33\textwidth}
        \centering
        \includegraphics[width=\linewidth]{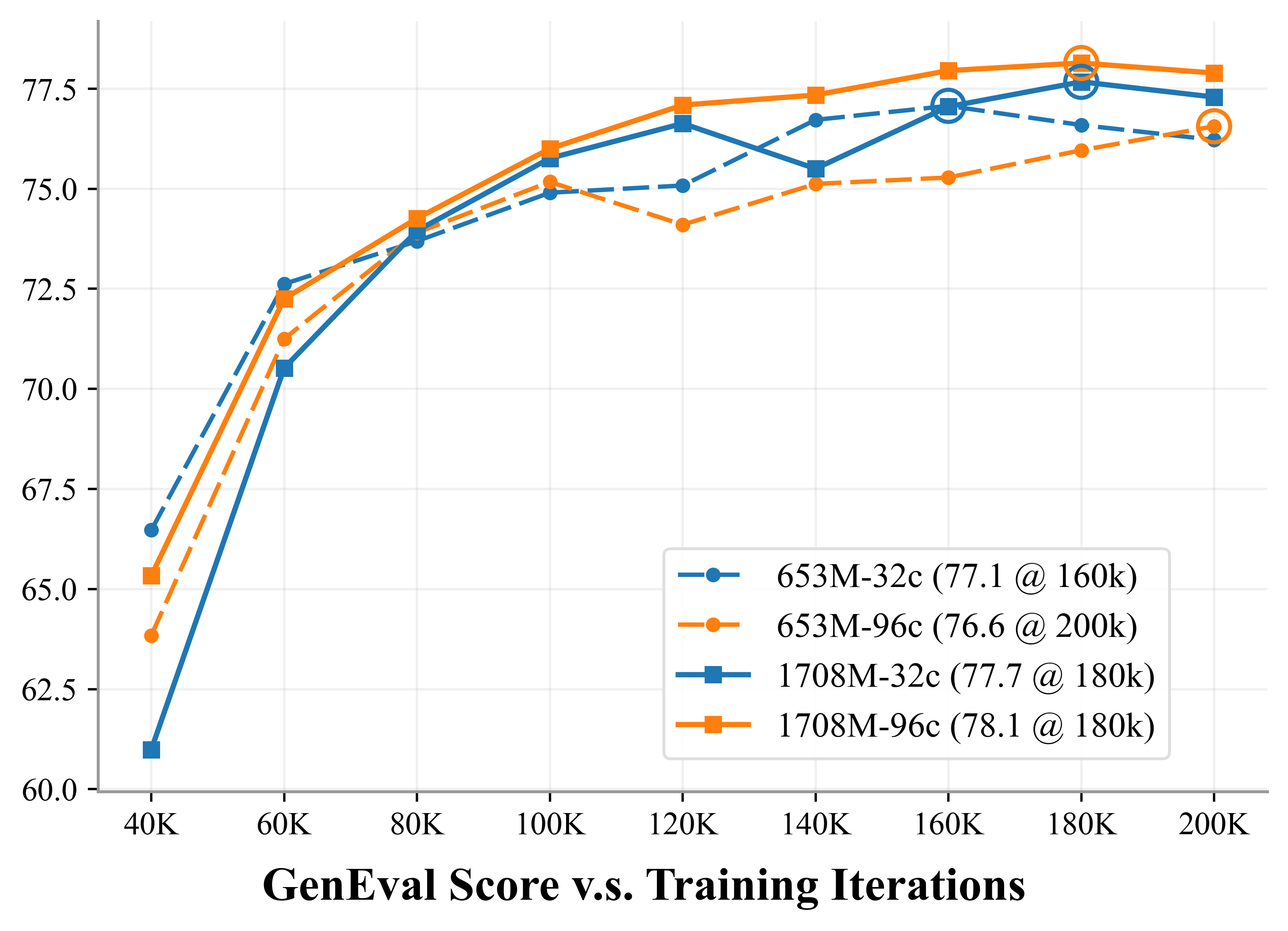}
        \vspace{-2mm}
        \label{fig:scaling-geneval}
    \end{minipage}
    \hfill
    \begin{minipage}[b]{0.33\textwidth}
        \centering
        \includegraphics[width=\linewidth]{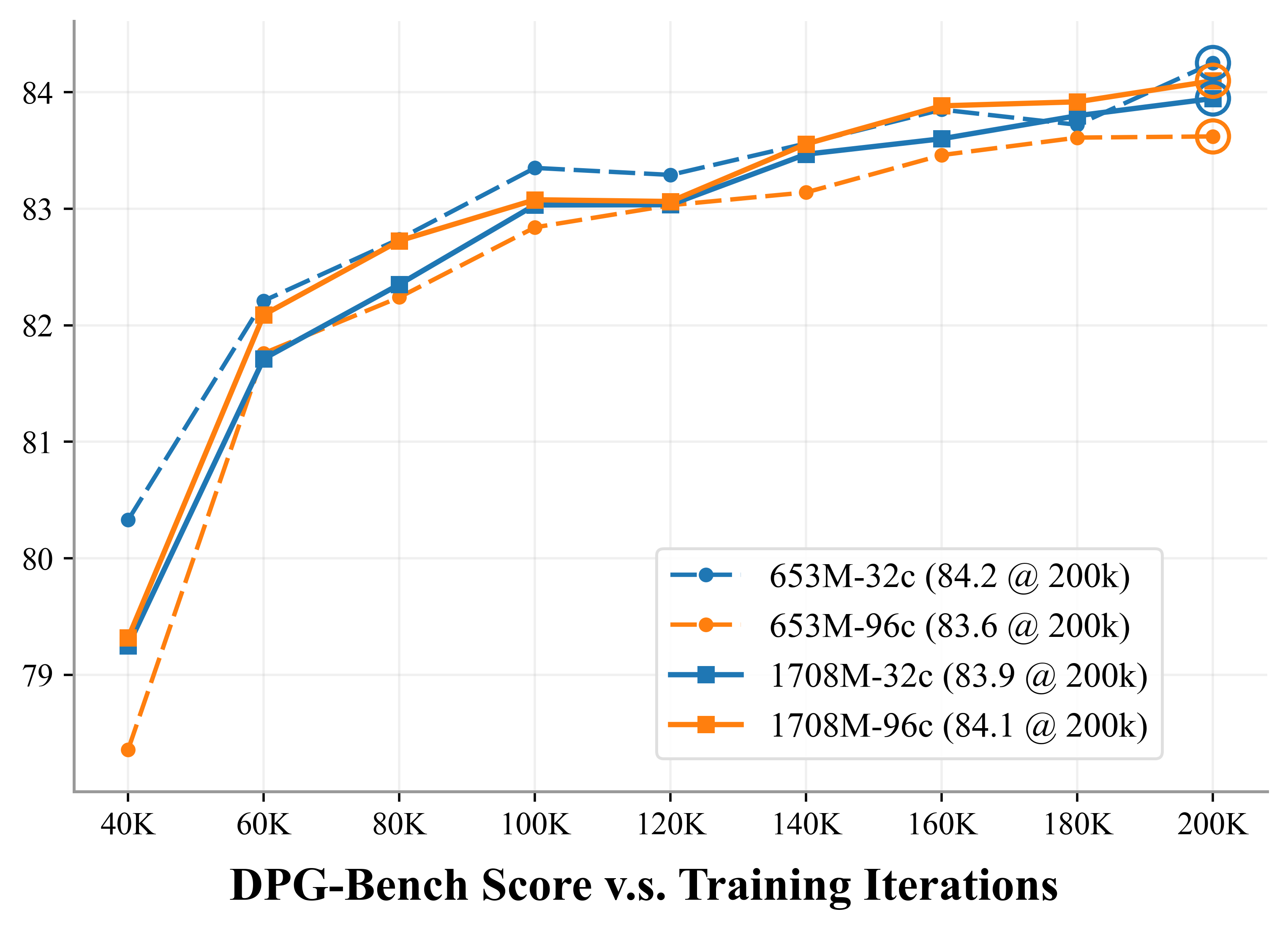}
        \vspace{-2mm}
        \label{fig:scaling-dpg}
    \end{minipage}
    \hfill
    \begin{minipage}[b]{0.33\textwidth}
        \centering
        \includegraphics[width=\linewidth]{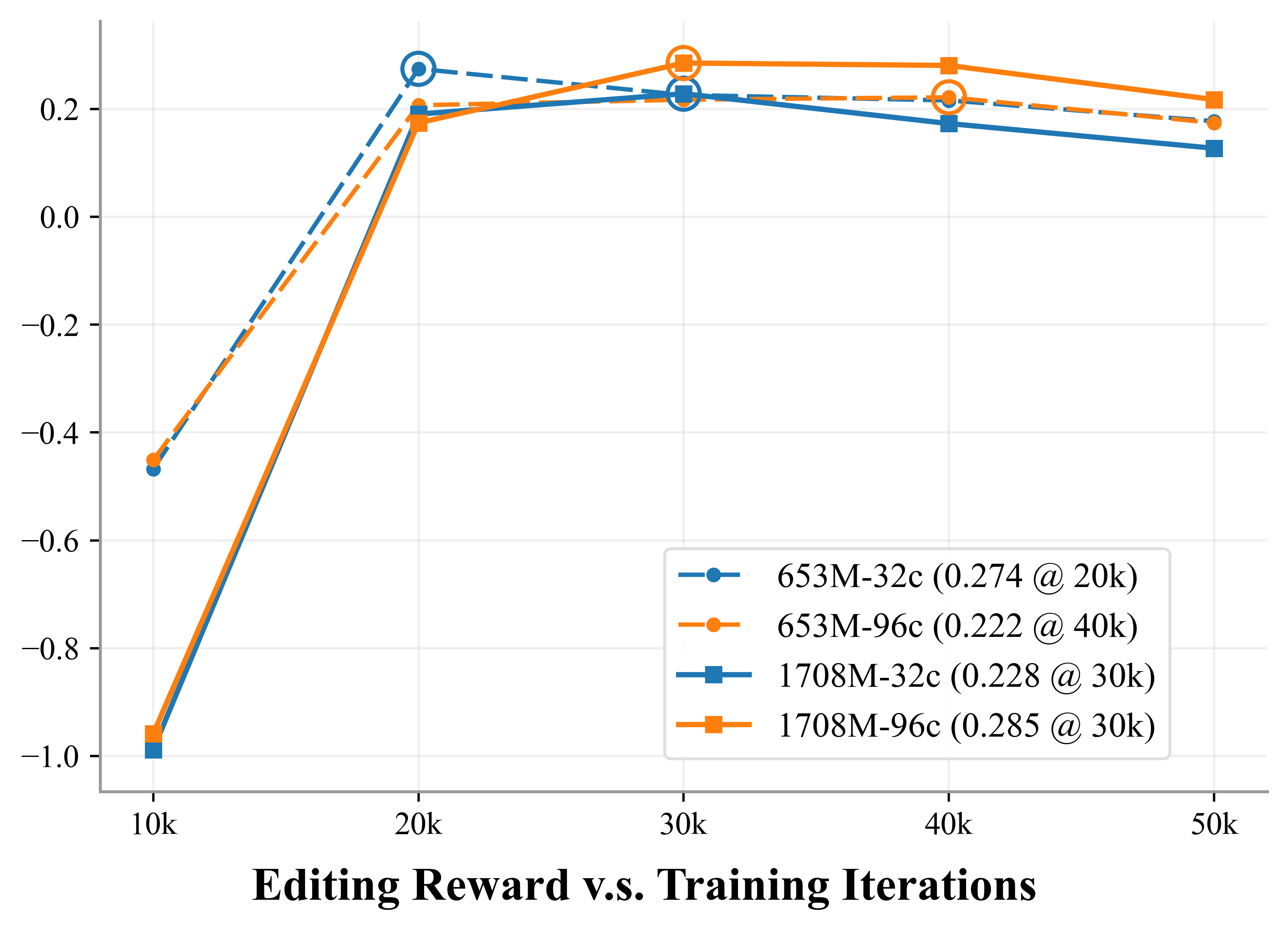}
        \vspace{-2mm}
        \label{fig:scaling-editing}
    \end{minipage}
    \vspace{-5mm}
    \caption{Scaling behavior of 653M (dashed) and 1708M (solid) models under different \ours\ channel sizes (32c/96c) on (a) GenEval, (b) DPG-Bench, and (c) Editing Reward.}
    \label{fig:scaling-two-benchmarks}
\end{figure*}

To assess whether the 96-channel variant offers a higher performance ceiling, we scale the generation backbone from Qwen-0.5B to Qwen-1.5B~\citep{qwen} under a fixed training setting, guided by the scaling observations in~\cite{esser2024scaling}.
This revealed distinct scaling behaviors as shown in \Cref{fig:scaling-two-benchmarks}: $\ours_{96c}$ exhibits consistent improvements across all tasks, with GenEval rising from 76.56 to 78.14, DPG-Bench from 83.62 to 84.09, and Editing Reward increasing significantly from 0.222 to 0.285. In contrast, $\ours_{32c}$ demonstrates diminishing performance, showing only marginal gains on GenEval (77.07 to 77.67) and slight degradation on both DPG-Bench (84.25 \vs 84.10) and Editing Reward (0.274 \vs 0.228). These results indicate that higher-channel latent spaces possess superior scaling properties and higher upper bounds when paired with a larger generative model. Investigating the correspondence between high-channel latent spaces and large-scale generation backbones represents a promising direction for future research.

\begin{figure*}[!t]
\begin{center}
\includegraphics[width=1\linewidth]{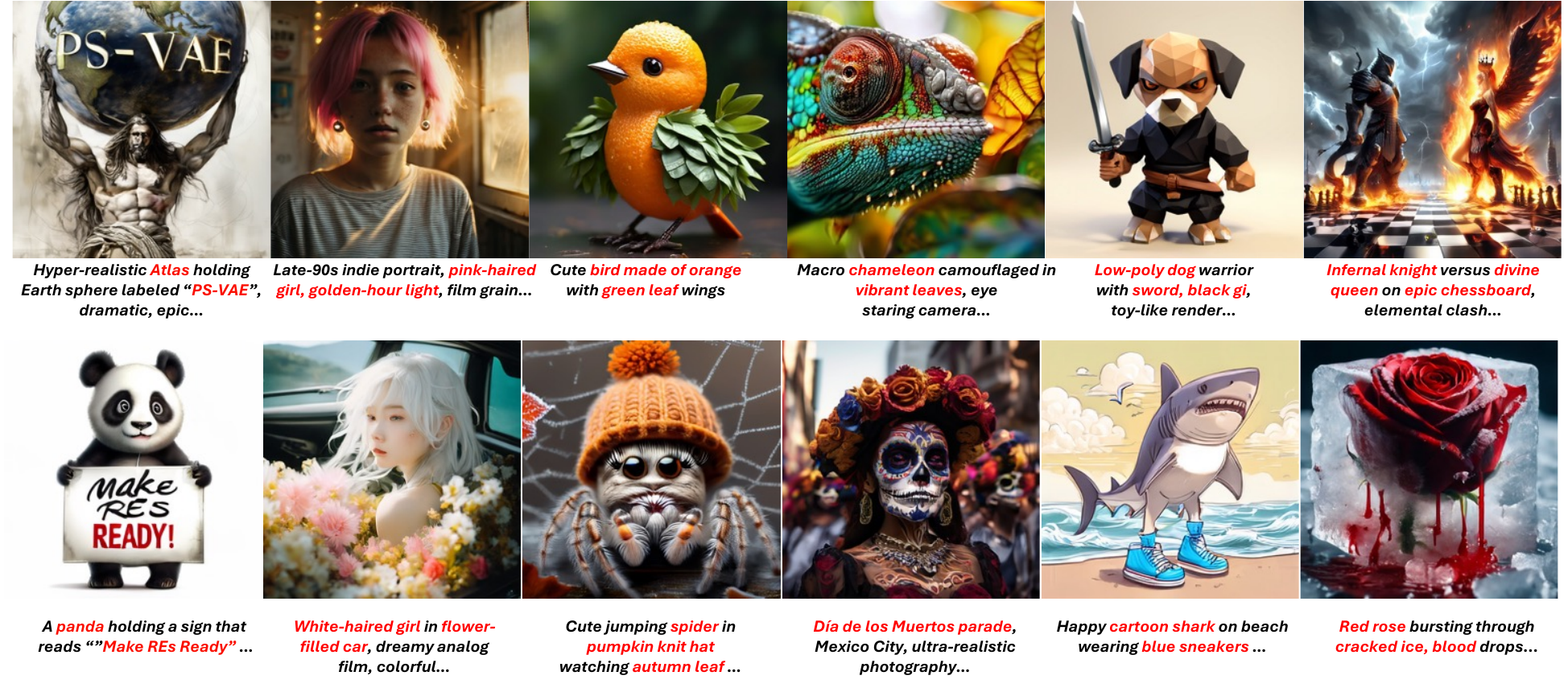}
\end{center}
\caption{\textbf{Text-to-image generation examples using $\ours_{96c}$. }
Despite being trained only at $256\times256$ resolution, the semantically structured and detail-preserving latent space enables the generator to accurately follow complex text prompts, yielding images with correct structures, fine-grained textures, precise text rendering, realistic portraits, and flexible compositions of abstract concepts. 
Prompts are simplified for visualization; full prompts can be found in the \supp.}\label{fig:t2i_example}
\vspace{-4mm}
\end{figure*}

Finally, we fine-tune Qwen-3B~\citep{qwen} with $\ours_{96c}$ for 600k iterations, including 200k iterations under the same training setting and an additional 400k iterations on a high-quality internal dataset. Benefiting from the strong semantic alignment and high-fidelity reconstruction of our latent space, the generator produces images with accurate text rendering, high-quality portraits, and flexible compositions of complex concepts (see \Cref{fig:t2i_example}). Notably, it achieves realistic textures and prompt-following capabilities while trained solely at $256 \times 256$ resolution. Extending to higher resolutions will further amplify these capabilities, which we leave as a direction for future exploration.

\subsection{\ours with SigLIP2: A Unified Encoder for Understanding and Generation}

While SigLIP~\cite{siglip, siglipv2} is widely adopted as a vision encoder for multimodal understanding, we investigate whether our method enables SigLIP2 to function as a unified encoder for both understanding and generation. Specifically, we utilize the SigLIP2-so400m/14 encoder from Bagel~\cite{deng2025bagel}, training it under settings identical to DINOv2 for fair comparison. The only modification is the weighting ratio between the semantic loss $L_s$ and the pixel reconstruction loss $L_p$, which is adjusted from $0.1{:}1$ to $0.05{:}1$. We observe that under equal loss weighting, the reconstruction quality of SigLIP2 saturates earlier than DINOv2, likely due to its more high-level and abstract representations.

\noindent \textbf{Generation Performance.}  
As shown in Table~\ref{tab:siglip2}, $\ours_{96c}(\texttt{DINOv2})$ and $\ours_{96c}(\texttt{SigLIP2})$ achieve comparable reconstruction quality with similar rFID, PSNR, LPIPS, and SSIM scores. In generation tasks, $\ours_{96c}(\texttt{SigLIP2})$ attains a slightly higher GenEval score, while $\ours_{96c}(\texttt{DINOv2})$ performs better on DPG-Bench and Editing Reward. Overall, both encoders demonstrate similar generation capabilities, confirming the robust transferability of our method across different pretrained foundation models.

\noindent \textbf{Understanding Performance.}  
To evaluate whether optimizing for reconstruction degrades semantic representations, we integrate our fine-tuned encoder into the original Bagel~\cite{deng2025bagel} pipeline, replacing the original encoder while keeping the LLM parameters frozen. On standard benchmarks, we observe negligible performance degradation in the zero-shot settings. For example, MME-P~\cite{fu2023mme} scores decrease slightly from 1685 to 1652, and VBench~\cite{liu2024mmbench} drops marginally from 85.0 to 84.7. These results validate that our proposed \ours preserves the core semantic capabilities of the original SigLIP2. It is worth noting that these results are obtained without any fine-tuning of the LLM parameters. We hypothesize that jointly training with the LLM could further enable the model to surpass the original baseline, which is left as a future extension, as the fine-tuned encoder preserves all visual information while maintaining a well-structured semantic representation. Overall, these results suggest that SigLIP2, optimized via \ours, holds the potential as a unified encoder for future understanding and generation architectures.

\begin{table*}[!t]
\centering
\caption{Comparison of reconstruction and generation performance between DINOv2 and SigLIP2.}
\label{tab:siglip2}
\resizebox{\columnwidth}{!}{
\begin{tabular}{lccccccccc}
\toprule
\textbf{Method} 
& \textbf{rFID}~(↓) 
& \textbf{PSNR}~(↑) 
& \textbf{LPIPS}~(↓) 
& \textbf{SSIM}~(↑) 
& \textbf{GenEval}~(↑)  
& \textbf{DPG-Bench}~(↑) 
& \textbf{Editing Reward}~(↑)  \\
\midrule

\textbf{$\ours_{96c}(\texttt{DINOv2-B})$} 
& 0.203 
& 28.79 
& 0.085 
& 0.817 
& 76.56
& 83.62 
& 0.222 \\

\textbf{$\ours_{96c}(\texttt{SigLIP2-so400m/14})$} 
& 0.222 
& 28.14
& 0.096 
& 0.795
& 77.14 
& 83.33
& 0.183  \\

\bottomrule
\end{tabular}}
\end{table*}

%% file: sec/5_ablation.tex
\section{Ablation Study} 

\subsection{Evolution from \texttt{RAE} to \ours}
We conduct a comprehensive ablation study to analyze how reconstruction and generation performance evolve as we progressively extend RAE to S-VAE and finally to \ours. The results are summarized in \Cref{tab:rae_to_our}, where MAR-VAE~\cite{li2024autoregressive} is included as a performance reference. 

\noindent \textbf{From \texttt{RAE} to \texttt{S-VAE}:}
\texttt{S-VAE} is mainly trained with the semantic reconstruction loss $\mathcal{L}_S$ and the KL loss $\mathcal{L}_{KL}$ to compact and regularize the feature space. Compared to \texttt{RAE}, \texttt{S-VAE} yields substantial improvements in generation and editing performance (GenEval: 71.3$\to$73.7, DPG-Bench: 81.7$\to$83.6, Editing Reward: 0.06$\to$0.12; see~\Cref{tab:rae_to_our}), despite a marked degradation in reconstruction quality (PSNR: 19.2$\to$17.78, SSIM: 0.436$\to$0.390). This suggests that \texttt{RAE}'s primary limitation stems not from reconstruction fidelity, but from the off-manifold nature of its high-dimensional semantic features, which \texttt{S-VAE} effectively mitigates.

Besides, we also compare \texttt{S-VAE} with \texttt{MAR-VAE} \cite{li2024autoregressive}. Results indicate that while \texttt{S-VAE} trails \texttt{MAR-VAE} in pixel-level reconstruction, its robust semantic representations allow it to outperform \texttt{MAR-VAE} on the semantics-oriented DPG-Bench~\cite{dpg} and in instruction-based editing. However, \texttt{S-VAE} underperforms on GenEval~\cite{geneval}.
Since GenEval relies on object detectors that are sensitive to texture and structure, the degraded pixel fidelity of \texttt{S-VAE} prevents the generation model from learning realistic textures and fine-grained structural details, leading to detection failures. This underscores the necessity of accurate reconstruction in VAEs for learning fine-grained object details.

\noindent \textbf{The Proposed \ours:} By integrating fine-grained detail supervision while maintaining semantic structure, our \ours recovers high-frequency details without sacrificing semantic coherence. The 96-channel \ours achieves state-of-the-art reconstruction, surpassing \texttt{MAR-VAE} on GenEval (+0.9) while retaining the strong DPG-Bench performance of \texttt{S-VAE}. In instruction-based editing, this enhanced visual fidelity significantly improves image–edit consistency, nearly doubling the editing reward (0.12$\to$0.22).

\noindent \textbf{The Role of Semantic Structure \texttt{(P-VAE)}:} We isolate the specific contribution of the semantic structure of latent space by training a Pixel-VAE \texttt{(P-VAE)} using solely the pixel reconstruction objective ($\mathcal{L}_P$). Linear probing indicates that its semantic quality regresses to the level of \texttt{MAR-VAE}. Consequently, DPG performance declines (83.6$\to$82.6) and editing reward drops sharply (0.22$\to$0.04). This confirms that explicit semantic regularization within the latent space is indispensable for text alignment and instruction following in both generation and editing tasks.

\begin{table}[t]
\centering
\caption{\textbf{Evolving the Representation Autoencoder (RAE) into our Pixel–Semantic Variational Autoencoder (PS-VAE).}
 Semantic regularization is essential for alleviating off-manifold behavior and improving generation and instruction-based editing performance (RAE $\rightarrow$ S-VAE), while enriching pixel details without disrupting semantic structure yields PS-VAE with the best overall performance.
In contrast, a pixel-only VAE (P-VAE) trained with only pixel reconstruction loss exhibits degraded semantic quality, as indicated by linear probing, along with inferior DPG-Bench and editing performance, despite achieving better pixel-level reconstruction. This highlights the importance of semantic structure in the latent space.
}

\resizebox{\columnwidth}{!}{
\begin{tabular}{l|ccccc|ccc}
\toprule
\textbf{Method} & \textbf{rFID}~(↓) & \textbf{PSNR}~(↑) & \textbf{LPIPS}~(↓) & \textbf{SSIM}~(↑) & \textbf{Linear Probe}~(↑)  & \textbf{GenEval}~(↑)  & \textbf{DPG}~(↑) & \textbf{Editing Reward}~(↑) \\
\midrule
MAR-VAE 
& 0.534  & 26.18  & 0.135  & 0.715  & 5.4/15.1 & 75.7 & 83.2  & 0.06 \\
\midrule
RAE   & 0.619  & 19.20  & 0.254  & 0.436  & 83.0/96.6 & 71.3   & 81.7 & 0.06  \\
S-VAE & 1.407  & 17.78  & 0.296  & 0.390  & 81.1/95.7 & 73.7   & 83.6  & 0.12 \\
P-VAE & 0.398  & 29.81  & 0.073  & 0.850  & 11.8/26.7 & 75.2   & 82.1   &0.04 \\
\midrule
\textbf{\ours}   & 0.203 & 28.79 & 0.085 & 0.817  & 79.5/94.8 & 76.6   & 83.6 & 0.22  \\
\bottomrule
\end{tabular}
}
\label{tab:rae_to_our}
\end{table}

\begin{figure*}[!t]
\vspace{-2mm}
\centering

\begin{minipage}[t]{0.34\textwidth}
\vspace{0pt}
\centering
\resizebox{\linewidth}{!}{%
\setlength{\tabcolsep}{1pt}
\begin{tabular}{lcccc}
\toprule
\textbf{\#C} & \textbf{rFID}~(↓) & \textbf{PSNR}~(↑) & \textbf{LPIPS}~(↓) & \textbf{SSIM}~(↑) \\
\midrule
32  & 0.584 & 24.53 & 0.168 & 0.662 \\
48  & 0.475 & 25.43 & 0.147 & 0.697 \\
64  & 0.423 & 26.65 & 0.124 & 0.744 \\
80  & 0.292 & 27.38 & 0.107 & 0.772 \\
96  & 0.203 & 28.79 & 0.085 & 0.817 \\
112 & 0.159 & \textbf{30.51} & \textbf{0.064} & \textbf{0.865} \\
256 & \textbf{0.156} & 30.30 & 0.065 & 0.860 \\
\bottomrule
\end{tabular}}
\end{minipage}\hfill
\begin{minipage}[t]{0.32\textwidth}
\vspace{0pt}
\centering
\includegraphics[width=\linewidth]{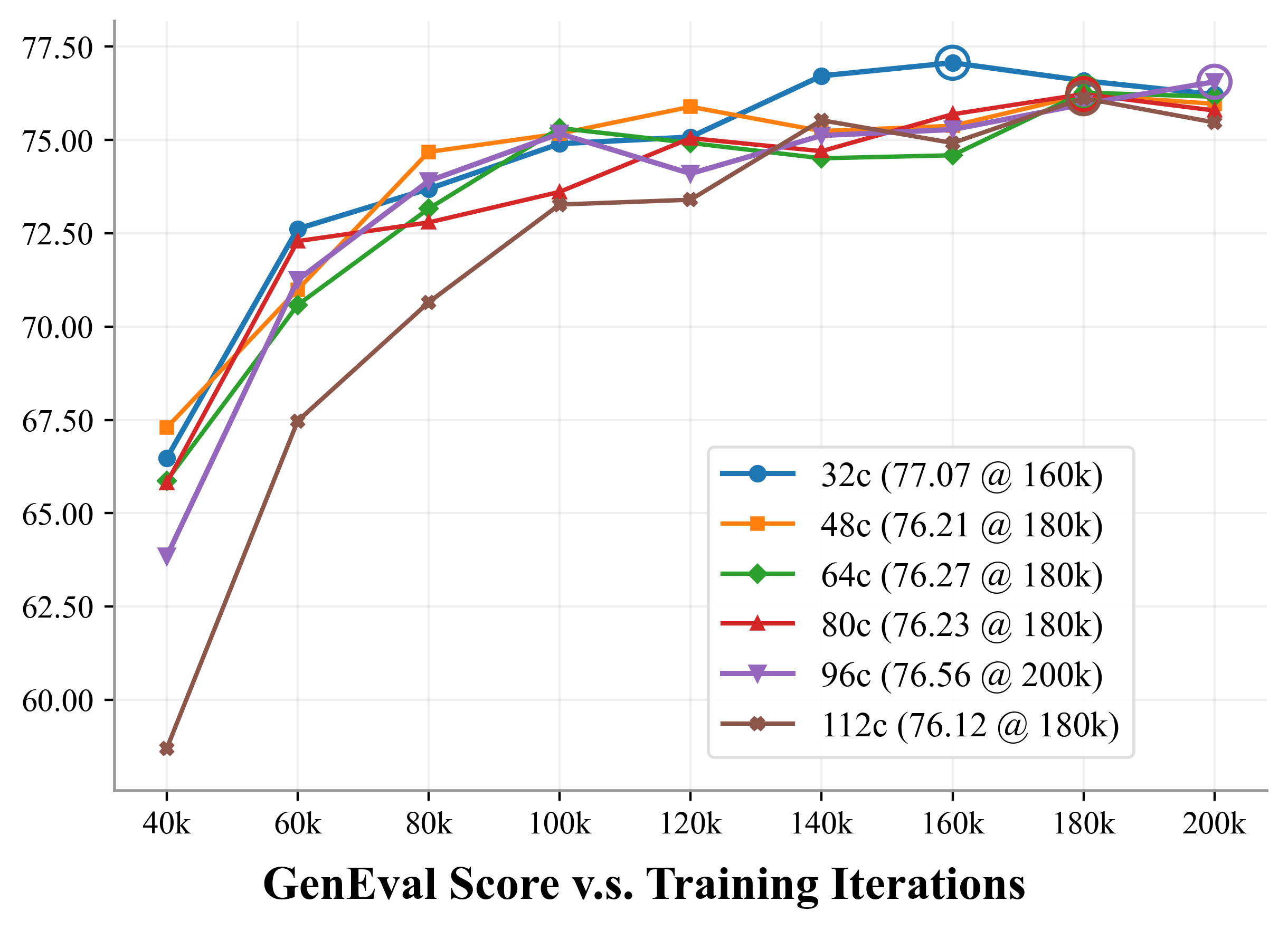}
\end{minipage}\hfill
\begin{minipage}[t]{0.32\textwidth}
\vspace{0pt}
\centering
\includegraphics[width=\linewidth]{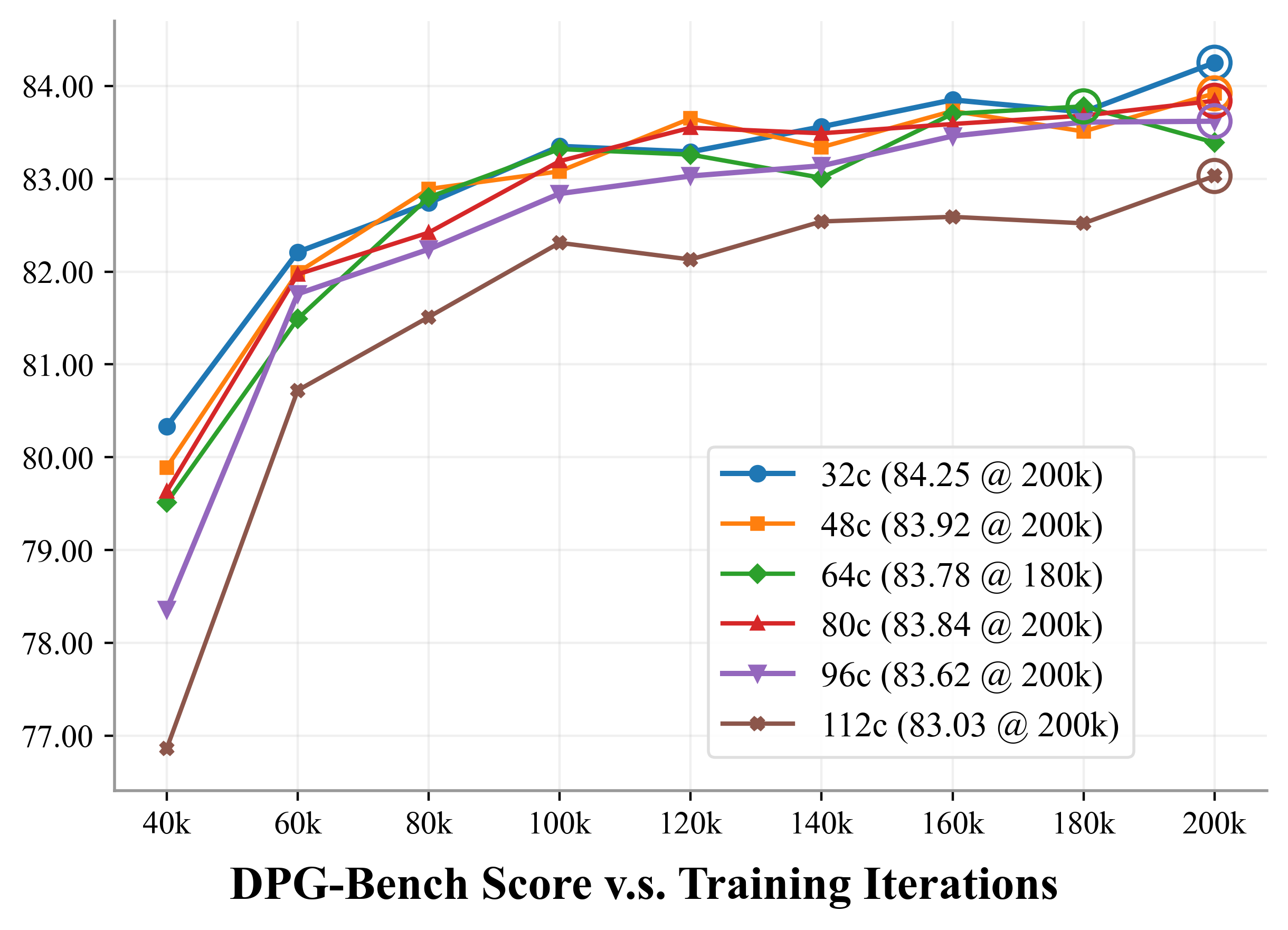}
\end{minipage}

\vspace{-3mm}
\caption{\textbf{Channel ablation of \ours.}
\textit{Left:} reconstruction metrics versus channel dimensionality.
\textit{Middle/Right:} convergence curves on GenEval and DPG-Bench.
As channel dimensionality increases, convergence becomes slightly slower; GenEval and DPG-Bench remain stable from 32c to 96c, while DPG-Bench drops beyond 96c, suggesting an intrinsic latent dimensionality of approximately $\sim$96 channels under our training setup. Further increasing the channel dimension primarily captures high-frequency details, which may consume model capacity and interfere with semantic learning.}
\label{fig:channel_ab}
\vspace{-4mm}
\end{figure*}

\subsection{Ablation on Latent Channel of \ours}
To find the optimal latent space dimensionality of \ours, we conduct a grid search (from 32 to 256) on the channel number. As shown in \Cref{fig:channel_ab}, reconstruction performance saturates at 112 channels, with further increases in width yielding negligible gains. As for generation performance, increasing latent dimensionality slows convergence (see \Cref{fig:channel_ab}). However, final performance remains comparable between 32 and 96 channels on both GenEval and DPG-Bench. A turning point occurs at 112 channels, where the DPG-Bench score drops noticeably by approximately 0.6 points. This observation suggests that, under our training setup, the intrinsic latent dimensionality required to jointly preserve semantic structure and pixel-level fidelity is around $\sim$96 channels.  Further increasing the channel capacity mainly introduces high-frequency details that consume additional model capacity, which might hinder semantic alignment, ultimately degrading generation performance.

\subsection{Ablation on Encoder and Decoder Architectures}

As shown in \Cref{tab:ae_arch}, projecting an unconstrained representation space into a compact, KL-regularized latent manifold~\cite{kingma2013auto} need non-trivial computational overhead. 
For example, a shallow 2-layer MLP is insufficient to preserve rich semantic features within a 96-channel latent space, as evidenced by a substantial drop in linear probing accuracy. 
This suggests that limited mapping capacity prevents the semantic VAE from fully capturing the intrinsic dimensionality of the presentation feature, leaving the effective intrinsic dimensionality below 96. Under the KL regularization constraint, this further induces posterior collapse, which ultimately degrades generation performance.

We also explore an architecturally symmetric design that directly feeds the reconstructed semantic features into the pixel decoder for final image reconstruction.
While this approach improves reconstruction quality, we observe a degradation in GenEval performance. We hypothesize that this stems from gradient interference, as both semantic and pixel reconstruction objectives are backpropagated through a shared Transformer pathway. Alternatively, this architecture may require a distinct re-balancing of loss weights to function effectively. We leave a deeper investigation of these dynamics to future work.

\begin{table*}[t]
\centering
\caption{\textbf{Comparison of reconstruction and generation performance across different PS-VAE design variants.}
{2L-MLP S-VAE} denotes an S-VAE in which both the semantic encoder and the semantic decoder are implemented as 2-layer MLPs.
$D_{\text{Pixel}}$ on $D_{\text{Semantic}}$ indicates that the pixel decoder is attached to the semantic decoder features during the final detail-enrichment training stage.}

\resizebox{\columnwidth}{!}{
\begin{tabular}{lccccccc}
\toprule
\textbf{Method} 
& \textbf{rFID}~(↓) 
& \textbf{PSNR}~(↑) 
& \textbf{LPIPS}~(↓) 
& \textbf{SSIM}~(↑) 
& \textbf{Linear}~(↑) 
& \textbf{GenEval}~(↑)  
& \textbf{DPG-Bench}~(↑) \\
\midrule
\textbf{\ours}   
& 0.214  
& 28.63  
& 0.087 
& 0.813 
& 79.5 / 94.8 
& 76.6   
& 83.6 \\

2L-MLP S-VAE 
& 0.205
& 28.94 
& 0.082
& 0.812
& 44.5 / 64.5 
& 70.3 
& 83.1 \\

$D_{\text{Pixel}}$~on~$D_{\text{Semantic}}$
& 0.193
& 29.64
& 0.077
& 0.840
& 80.4 / 95.4
& 74.4 
& 83.6 \\
\bottomrule
\end{tabular}}
\label{tab:ae_arch}
\vspace{-3mm}
\end{table*}

\subsection{Directly Enriching High-Dimensional Features Fails}

\begin{wraptable}{r}{0.45\textwidth}  
\centering
\vspace{-5mm}
\caption{\textbf{High-dimensional enrichment causes shortcut reconstruction.}
RAE-HD greatly improves reconstruction metrics but harms generation, indicating loss of semantic structure and using shortcut reconstruction.}
\label{tab:enrich_on_high}
\resizebox{\linewidth}{!}{  
\begin{tabular}{l|ccccc}
\toprule
\textbf{Method} 
& \textbf{rFID}↓ 
& \textbf{PSNR}↑ 
& \textbf{LPIPS}↓ 
& \textbf{SSIM}↑ 
& \textbf{GenEval}↑ \\
\midrule

RAE 
& 0.619 
& 19.20 
& 0.254 
& 0.436 
& 71.3 \\
\midrule
RAE-HD
& 0.193
& 33.10
& 0.048
& 0.916 
& 60.2 \\

\bottomrule
\end{tabular}}
\vspace{-10pt}
\end{wraptable}

\begin{wrapfigure}{r}{0.45\textwidth}
\begin{center}
\vspace{-5mm}
\includegraphics[width=1\linewidth]{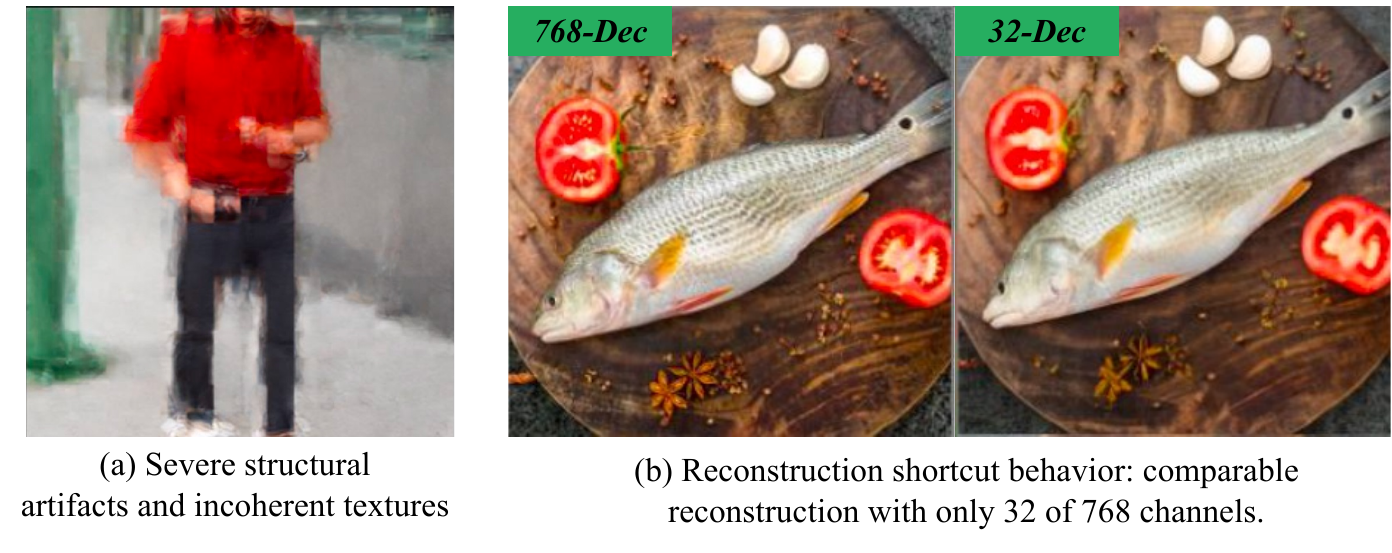}
\end{center}
\vspace{-3mm}
\caption{Directly enriching details in a high-dimensional space leads to severe generation artifacts (a). 
We further verify that this behavior arises from reconstruction shortcuts in high-dimensional feature spaces (b).}
\label{fig:rec_shortcut}
\vspace{-3mm}
\end{wrapfigure}

An alternative strategy to improve pixel reconstruction involves training the pixel decoder directly on the original high-dimensional feature space, while maintaining the semantic reconstruction loss with a frozen DINOv2 encoder. As shown in \Cref{tab:enrich_on_high}, while this approach yields rapid improvements in reconstruction quality, it leads to a significant degradation in generation performance. The generated images exhibit severe structural artifacts and incoherent textures(as shown in ~\Cref{fig:rec_shortcut}.a).

We attribute this failure to the inherent difficulty of constraining a high-dimensional latent space. Even with semantic-preserving losses, the model can exploit shortcut solutions by relying on a sparse subset of channels for reconstruction, without inducing meaningful changes in feature distances in high-dimensional spaces, where distance metrics tend to become less informative.
We verify this shortcut behavior by showing that retraining a pixel decoder using only the 32 selected channels (out of 768) with the largest deviations between the fine-tuned encoder and the frozen DINOv2 features is sufficient to achieve strong reconstruction performance (as shown in ~\Cref{fig:rec_shortcut}.b). This indicates that constraining detail enrichment to a compact, semantically regularized latent space is essential, thereby validating the core design of our \ours.

%% file: sec/6_conclusion.tex
\subsection{Conclusions}

In this work, we show that powerful representation encoders, despite their strong discriminative ability, are not directly suitable as generative spaces due to unconstrained feature distributions and insufficient image reconstruction fidelity.
Through systematic analysis, we identify off-manifold generation and poor reconstruction as the two key bottlenecks limiting their performance in text-to-image generation and instruction-based editing.
To address this, we propose a Pixel–Semantic VAE (\ours) that maps representation features and pixel details into a compact, KL-regularized latent space by properly finetuning pre-trained representation encoders under both pixel and semantic reconstruction objectives. 
As a result, \ours achieves state-of-the-art performance in reconstruction, generation, and image editing.
We believe this work offers a practical pathway toward unifying visual understanding and generation within a single encoder.